%% file: root.tex
\documentclass[letterpaper, 10 pt, conference]{ieeeconf}  %

\IEEEoverridecommandlockouts                              %

\usepackage{graphicx} %
\usepackage{amsmath} %
\usepackage{amssymb}  %

\usepackage[font=footnotesize]{caption}
\usepackage{balance}
\usepackage{nameref}
\usepackage{siunitx}
\usepackage{subcaption}
\usepackage{multirow}
\usepackage{booktabs}
\usepackage{tabularx}
\usepackage{tikz}
\usepackage{hyperref}
\usepackage{pgfplots}
\usepackage{tikzscale}
\usepackage{stmaryrd}
\pgfplotsset{compat=newest}
\newlength\figureheight
\newlength\figurewidth 

\usepackage[noadjust]{cite}
\usepackage{siunitx}

\usetikzlibrary{calc}

\usepackage{color}
\usepackage{import}
\newcommand{\executeiffilenewer}[3]{%
	\ifnum\pdfstrcmp%
		{\pdffilemoddate{#1}}%
		{\pdffilemoddate{#2}}%
		>0%
		{\immediate\write18{#3}}%
	\fi%
}

\title{\LARGE \bf
	Deep Object Tracking on Dynamic Occupancy Grid Maps Using RNNs
}

\author{Nico Engel$^{1}$, Stefan Hoermann$^{1}$, Philipp Henzler$^{1}$ and Klaus Dietmayer$^{1}$ 
\thanks{The authors are with:
\newline
$^{1}$ Institute of Measurement, Control, and Microtechnology, Ulm University, Germany
	{\tt\small \{firstname.lastname\}@uni-ulm.de} %
}%
}

\def\mysection#1#2{\section{#1}\label{sec:#2}}
\def\mysubsection#1#2{\subsection{#1}\label{sec:#2}}

\include{commands}

\usepackage[top=0.78in, bottom=0.8in,left=0.77in, right=0.77in ]{geometry}

\begin{document}

\bstctlcite{IEEEexample:BSTcontrol} 

\def\cred{\textcolor{red}}
\def\cblue{\textcolor{blue}}
\def\cgreen{\textcolor{green}}

\def\etal{et\:al.\ }
\def\DOGMA{DOGMa}

\def\GMcell{c}
\def\GMtimestep{t}
\def\GMnuminliers{n}

\def\GMchannels{\Omega}
\def\GMwidth{W}
\def\GMheight{H}

\def\Anchor{\alpha}
\def\ANCnumshapes{C_{s}}
\def\ANCnumorientations{C_{\ANCori}}
\def\ANCnumanchors{C_{\Anchor}}
\def\ANCnumlengths{C_{\ANClength}}
\def\ANCnumwidths{C_{\ANCwidth}}
\def\ANCTolerance{\delta}

\def\OBJwidth{w}
\def\OBJlength{l}
\def\OBJori{\phi}
\def\OBJaspect{a}

\def\ANCwidth{\OBJwidth}
\def\ANClength{\OBJlength}
\def\ANCori{\OBJori}
\def\ANCapsect{\OBJaspect}
\def\ANCscale{\ANClength}

\def\ANCdWrel{\Delta{\ANCwidth}}
\def\ANCdLrel{\Delta{\ANClength}}
\def\ANCdOriRel{\Delta{\ANCori}}

\def\Occupied{\mathrm{O}}
\def\MassFree{M_\mathrm{F}}
\def\MassOcc{M_\mathrm{O}}
\def\East{\mathrm{E}}
\def\North{\mathrm{N}}

\def\static{\mathrm{s}}
\def\dynamic{\mathrm{d}}

\def\IoUthreshold{\gamma}

\def\Netout{\hat{y}}
\def\Label{y}

\def\NetoutStatic{\Netout^{(\static)}}
\def\NetoutIoU{\Netout^{(\mathrm{IoU})}}
\def\NetoutDW{\Netout^{(\ANCdWrel)}}
\def\NetoutDL{\Netout^{(\ANCdLrel)}}
\def\NetoutDOri{\Netout^{(\ANCdOriRel)}}

\def\LabelIoU{\Label^{(\mathrm{IoU})}}
\def\LabelDW{\Label^{(\ANCdWrel)}}
\def\LabelDL{\Label^{(\ANCdLrel)}}
\def\LabelDOri{\Label^{(\ANCdOriRel)}}

\def\NetoutIoUMax{\hat{\mathbf{A}}}
\def\LabelIoUMax{\mathbf{A}}

\def\FocusingParam{f}
\def\GainParam{\lambda_\mathrm{I}}

\maketitle
\thispagestyle{empty}
\pagestyle{empty}

\input{doc/abstract}

\input{doc/sec_1_introduction}

\input{doc/sec_2_related_work}

\input{doc/sec_system_architecture}

\input{doc/sec_generating_labels_short}

\input{doc/sec_loss_function}

\input{doc/sec_results}
\input{doc/sec_conclusion}
\bibliographystyle{IEEEtran}
\balance
\bibliography{IEEEtranControl,SH} %

\end{document}

%% file: commands.tex
\usepackage{soul}
\usepackage{amsmath}
\usepackage{amssymb}
\usepackage{microtype}
\usepackage{wrapfig}
\usepackage{todonotes}
\usepackage{graphicx}

\newcommand{\eg}{e.\,g.\ }
\newcommand{\ie}{i.\,e.\ }
\newcommand{\etal}{et~al.\ }

\newcommand{\refFig}[1]{Fig.~\ref{fig:#1}}

\soulregister\ref7
\soulregister\cite7
\soulregister\refFig7
\soulregister\cite7
\soulregister\ref7
\soulregister\pageref7
\soulregister\shortcite7
\soulregister\eg0
\soulregister\ie0
\soulregister\etal0

\DeclareGraphicsExtensions{.png,.jpg,.pdf,.ai,.psd}

%% file: doc/abstract.tex
\begin{abstract}

The comprehensive representation and understanding of the driving environment is crucial to improve the safety and reliability of autonomous vehicles. 
In this paper, we present a new approach to establish an environment model containing a segmentation between static and dynamic background and parametric modeled objects with shape, position and orientation. Multiple laser scanners are fused into a dynamic occupancy grid map resulting in a $360^{\circ}$ perception of the environment. A single-stage deep convolutional neural network is combined with a recurrent neural network, which takes a time series of the occupancy grid map as input and tracks cell states and its corresponding object hypotheses. The labels for training are created unsupervised with an automatic label generation algorithm. 

The proposed methods are evaluated in real-world experiments in complex inner city scenarios using the aforementioned $360^{\circ}$ laser perception. The results show a better object detection accuracy in comparison with our old approach as well as an AUC score of $0.946$ for the dynamic and static segmentation. Furthermore, we gain an improved detection for occluded objects and a more consistent size estimation due to the usage of time series as input and the memory about previous states introduced by the recurrent neural network.

\end{abstract}

%% file: doc/sec_1_introduction.tex
\mysection{Introduction}{sec:introduction}
In autonomous driving applications, one of the key challenges for almost every software module, e.g. trajectory planning, decision making or behavior planning, lies in the perception and the precise understanding of the environment \cite{kunz2015}. Grid maps are widely used to represent the environment by dividing the space into a finite amount of equal cells, which are independent, discretized regions of the surroundings. Each cell constitutes the state of the corresponding area (e.g. free, occupied, dynamic, static) \cite{elfes1990occupancy}. Besides grid maps that are based on an object-model-free representation, object-model-based tracking can be used to model the environment as well. For this, an object is usually described by its size and its pose, but dynamics, existence probabilities and other relevant properties can also be specified. The goal is to find and extract moving objects, usually by observing a distinguishable shape, and track the movement by associating measurements from one time step to another.

Object-model-free grid maps on the other hand, fuse raw sensor data from multiple sensors into one environment representation and neglect the detection of objects. Instead, the occupancy probability of each cell is estimated. Therefore, circumventing the association problem that arises when a decision has to be made to assign a measurement to a specific object. Furthermore, this approach allows for a $360^{\circ}$ birds eye view of the environment.
One downside of the object-model-free representation is the assumption of independence of single cells for efficient computation. This causes the borders of stationary objects, e.g., walls or static cars, to have false velocity estimates. Simple clustering algorithms usually yield bad results. Convolutional neural networks (CNNs) and recurrent neural networks (RNNs) can be trained to exploit context and sequential data and essentially overcome this problem. 

Based on our previous work \cite{Hoermann18Detection}, \cite{Hoermann17Prediction}, we propose a neural network with a convolutional Long-Short-Term-Memory (LSTM) Cell \cite{DBLP:journals/corr/ShiCWYWW15} embedded in a single-stage convolutional neural network. 
As input, we use a Bayesian filtered, object-model-free dynamic occupancy grid map (DOGMa)  \cite{NUSS2016}, that contains the Dempster-Shafer masses for free and occupied cells, as well as the velocities and their associated variances. In \cite{Hoermann18Detection} we detected objects and estimated the corresponding poses on a DOGMa. Since erroneous object hypotheses, such as missed detections or objects resulting from clutter, can lead to devastating consequences and fatal accidents, the ultimate goal is to have a comprehensive and profound understanding of the environment. Thus, we try to exploit the advantages of both the object-model-free DOGMa, which provides a complete context representation without making assumptions about semantics, and the object-based tracking. Our approach uses a Bayesian filtered perception at a first stage with a subsequent neural network and yields an extensive environment representation with segmentation of dynamic and static regions, as well as extracted, parameterized objects. Furthermore, we make use of time sequences with a LSTM, which tracks the cell states over time. 
Consequently, we keep all available information by avoiding early abstraction and complementing the advantages of model-free and model-based environment representations.

The remainder of this paper is structured as follows: Section \ref{sec:related_work} reviews related work in the area of dynamic occupancy grid maps, segmentation methods, object detection and deep learning approaches for grid maps. Our system architecture is presented in Section \ref{sec:system_architecture}. This includes the network architecture, an explanation of our RNN and its output, as well as a brief overview of our grid map.
The automatic label generation algorithm, that is used for creating our training dataset is introduced in Section \ref{sec:label_generation}.
The training of our neural network is described in Section \ref{sec:training}, including the definition of our loss function and our dataset. 
The results of our experiment are shown in Section \ref{sec:results} and an evaluation is carried out to compare the results with our previous contribution.
Finally, a conclusion and possible future work is given in Section \ref{sec:conclusions}.

%% file: doc/sec_2_related_work.tex
\mysection{Related Work}{related_work}
The adequate representation of the driving environment is crucial for all automated driving systems. For this, many information sources are usually combined into a comprehensive environment model, which is a simplified abstraction of the real world. A detailed overview of different representation types can be found, e.g. in \cite{schreier2016compact}.
Extensive research has been conducted on grid maps, which were first introduced by Elfes \cite{elfes1990occupancy}, \cite{Elfes1989} in the 1990s. Since then, 2D and 3D  representations were developed \cite{wurm2010octomap}, \cite{Fankhauser2016GridMapLibrary}, utilizing the advantages of model-free representations. Nuss \etal \cite{NUSS2016} proposed a Bayesian filtered, dynamic occupancy grid map, that we use in this contribution. The classification and tracking of dynamic cells can be performed probabilistically, e.g. with an unscented Kalman-Filter \cite{schreier2014grid}, or data driven, e.g. with a fully convolutional neural network \cite{Piewak2017GMObjects}, where every pixel is classified with the help of clustering algorithms. We however, obtain the same segmentation with additional object predictions determined by shape, position and orientation directly from the input DOGMa.  

A common approach for object tracking is to fit boxes into raw measurements and track the shapes. However, these methods mainly use hand engineered features like L-shapes in laser \cite{kim2018shape}, \cite{Munz2009Lshapes} or radar measurements \cite{Roos2016RadarVehicleOriEstimation}. L-shapes are a popular choice, because they resemble the front and side of a car, which can be seen by almost every sensor. Nevertheless, this approach suffers from simplifications regarding the sensors and the environment and relies heavily on heuristics. Scheel \etal \cite{Scheel2017learnedRadarModel} proposed a method, where the measurement model is learned, thus circumventing the manual selection of shapes. But this approach, so far, focuses on cars only.

Data driven methods for object detection have shown great success in the past. Approaches with convolutional neural networks like Faster R-CNN \cite{FasterRCNN} or Mask R-CNN \cite{MaskRCNN} employ a two stage object detection, where first a region of interest (ROI) is obtained and then a classification is performed on the ROI. 
Gan \etal \cite{gan2015first} trained an end-to-end recurrent neural network in combination with a CNN to detect unknown objects from a video clip. In their work, the RNN outputs bounding boxes and fuses past predictions along with visual features obtained by a CNN. However, synthesized data is used for offline training. A similar approach was done by Dequaire \etal \cite{dequaire2016deep} with an end-to-end RNN for object tracking, where pixel states were inferred.
This is in contrast to our method, where we obtain bounding boxes for dynamic objects. Furthermore, we embed a Long-Short-Term-Memory (LSTM) Cell in our CNN structure for temporal filtering of the object hypotheses. LSTM Cells were first introduced by Hochreiter and Schmidhuber \cite{hochreiter1997long} in 1997 and are successfully applied for object detection, e.g. in \cite{ning2017spatially}, where Ning \etal predict and track the location of objects.

%% file: doc/sec_system_architecture.tex
\mysection{System Architecture}{system_architecture}

\begin{figure}[t]
	\centering
	\includegraphics[width=0.9\linewidth]{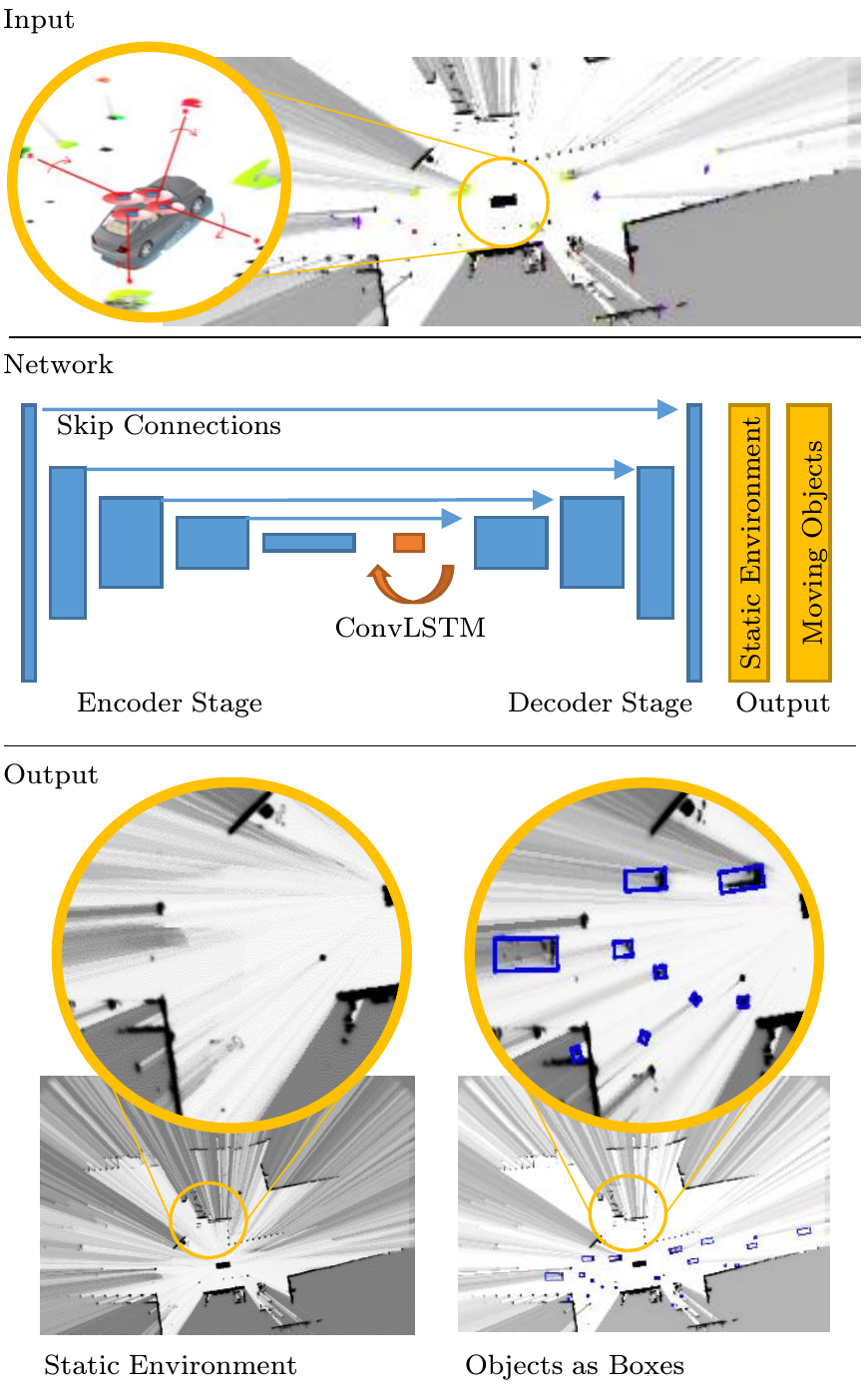}
	\caption{System Overview: The input is a dynamic occupancy grid map, shown as excerpt in the first row. 
		The second row depicts the employed neural network, containing a down sampling stage, a ConvLSTM cell, an upsampling stage and two heads as output.
		The resulting environment model is shown in the third row: The static environment as grid map where dynamic cells are removed by the network and moving objects as rotated rectangles.
	}
	\label{fig:system_architecture}
\end{figure}

The proposed system aims for an environment model containing static regions in a model-free representation and objects modeled as rectangles with position, shape and orientation.
An overview of the system architecture is given in Fig.~\ref{fig:system_architecture}.

The dynamic occupancy grid map (\DOGMA) from \cite{NUSS2016} serves as input to a neural network.
The \DOGMA\ can be seen as a full object-model-free environment representation based on Bayesian sensor fusion.
Occupancy and velocity estimates for grid cells $\GMcell$ result from a particle filter.
The cell channels $\GMchannels = \left\{ \MassOcc, \MassFree, v_\East, v_\North, \sigma^2_{v_\East}, \sigma^2_{v_\North}, \sigma^2_{v_\East, v_\North} \right\}$ denote the Dempster-Shafer \cite{dempster2008generalization} masses for occupancy and free space, the velocity pointing east and north, as well as the velocity variances and covariances, respectively.
In our grid illustrations, dark pixels refer to high occupancy probability, calculated with $P_\Occupied = 0.5\cdot\MassOcc + 0.5\cdot(1-\MassFree)$.
Occupancy probability at a grid cell $\GMcell$ and sequence time step $\GMtimestep$ is denoted by $P_\Occupied(\East, \North, \GMtimestep) := P_\Occupied(\GMcell, \GMtimestep)$.
The \DOGMA\ has width $\GMwidth$ and height $\GMheight$, providing data in  $\mathbb{R}^{\GMwidth \times \GMheight \times |\GMchannels|}$ for each time step.
A convolutional neural network transforms the object-model-free dynamic representation of the \DOGMA\ into a model-free representation of static regions and model-based representation of moving objects.
The employed neural network is based on a simple encoder-decoder structure with skip connections \cite{ronneberger2015u}. 
A ConvLSTM \cite{DBLP:journals/corr/ShiCWYWW15} cell is included upstream to the decoder stage with the intention to accumulate information from the input sequence.
Although the \DOGMA\ particle filter provides information about dynamics in the scene, long-term sequence information accumulation by recurrent network cells can help to compensate long-term occlusions and non-Gaussian noise which is hard to overcome with common grid fusion.
In addition, the convolutional LSTM cells are able to include context in sequential processing, while \DOGMA\ cells are assumed to be independent in the particle filter.
The \DOGMA\ side length is $\East=\North=901$ while step-wise downscaling is performed in the encoder to $301$, $101$, $51$ and $26$.
A $26\times26\times512$ tensor is fed to the LSTM cell with kernel size $5\times5$ and $512$ output channels. 
The result is upscaled using deconvolution mirroring the downscaling stages.

Parallel network heads form the output of our environment model.
The first output $\NetoutStatic$ is the occupancy probability of static regions provided in  $\mathbb{R}^{\GMwidth \times \GMheight \times 1}$.
The $4$ output tensors $\NetoutIoU$, $\NetoutDW$, $\NetoutDL$ and $\NetoutDOri$ yield object rectangles following the concept of anchors \cite{SSD}.
Anchors define a set of default boxes described by triples  $\left(\ANCwidth, \ANClength, \ANCori \right)$, including box width, length and orientation, respectively.
$\NetoutIoU$ provides a score in terms of the expected intersection-over-union (IoU), which is the Jaccard-Index, between a detected object and anchors.
Relative offsets $\ANCdWrel$, $\ANCdLrel$ and $\ANCdOriRel$ between detected object rectangle and default anchor box are given in the remaining output tensors.
We used the optimized anchors from \cite{Hoermann18Detection}, resulting in tensors $\NetoutIoU \in \mathbb{R}^{\GMwidth \times \GMheight \times 120}$, $\NetoutDW, \NetoutDL \in \mathbb{R}^{\GMwidth \times \GMheight \times 10}$ and $\NetoutDOri \in \mathbb{R}^{\GMwidth \times \GMheight \times 12}$.

%% file: doc/sec_generating_labels_short.tex
\mysection{Automatic Label Generation}{label_generation}

\begin{figure}[t]
	\centering
	\includegraphics[width=0.8\linewidth]{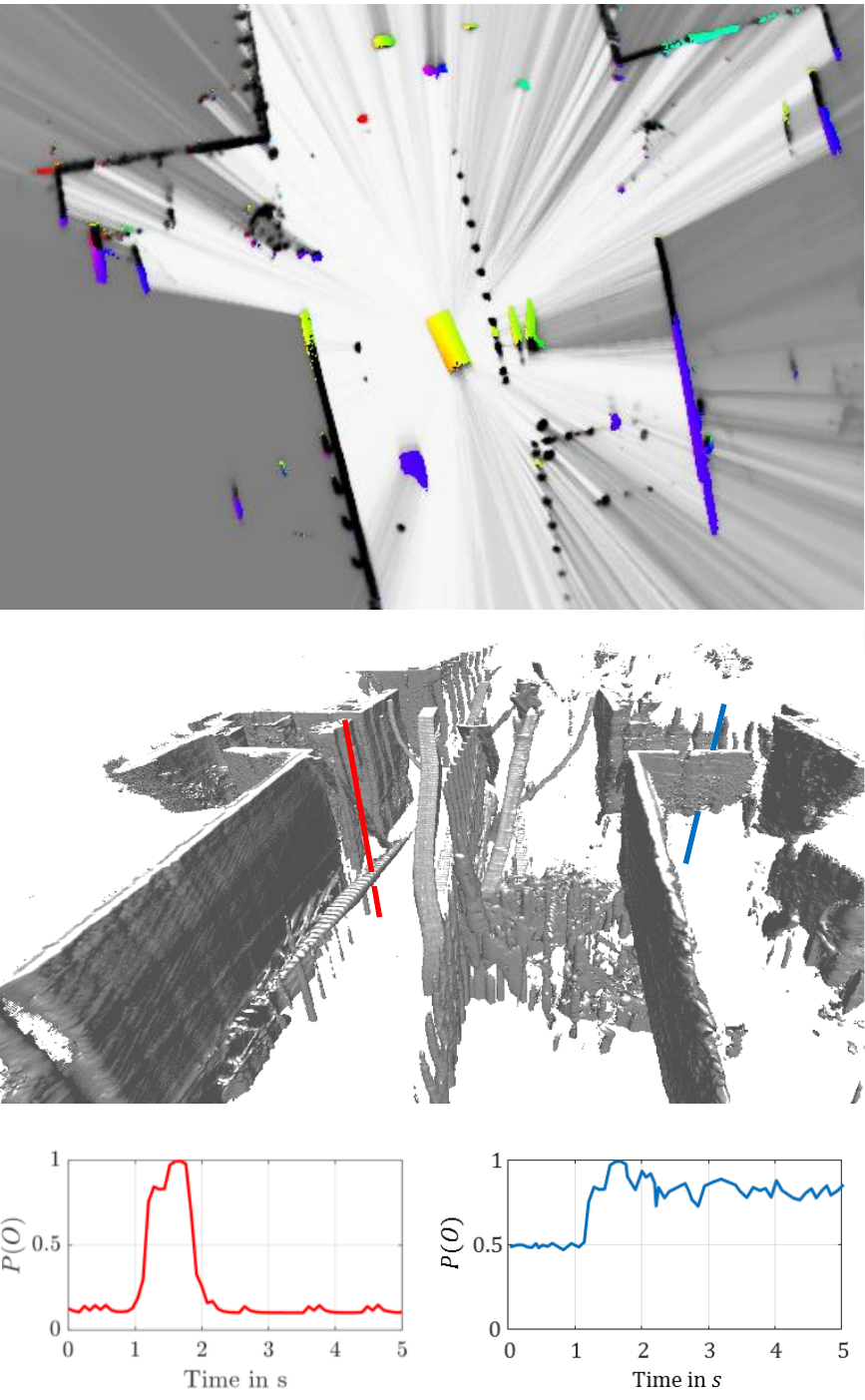}
	\caption{Offline sequence assessment: A single \DOGMA\ time step contains false velocity estimation, as illustrated by colored pixels in the DOGMa (top row). 
	In the 3D-graph (middle row), the grid map sequence is aligned using ego motion compensation.
	Occupancy probability of the two vertical lines (red and blue) is illustrated in the bottom row. 
	The red curve illustrates $P_\Occupied(t)$ caused by moving objects, while at the blue curve, static regions become uncovered causing false velocity estimates in the particle filter.
	}
	\label{fig:emags}
\end{figure}

Labels are generated fully automatically using offline sequence processing.
While real-time applications usually don't refine past estimates, for our goal the whole sequence can be used to generate accurate labeling data.
Instead of only relying on velocity estimates in the \DOGMA, occupancy movement is analyzed as illustrated in Fig. \ref{fig:emags}.
This enables us to overcome measurement noise and distinguish between actual moving objects and static objects with false velocity estimation.
False velocity estimation in grid cells is mainly caused when static regions enter the field of view or become visible from previous occlusion.
Therefore, we search for a pattern in the occupancy probability sequence of cells $P_\Occupied(\GMcell,\GMtimestep)$: a raise of $P_\Occupied$ followed by a fall.
The resulting classification is directly used to train $\NetoutStatic$.
For object labels, we first fit rectangles straight forward in dynamic regions, but refine their shape and orientation when the object was traced through the sequence.
Furthermore, unreasonable trajectories are removed from the labeling data.
A detailed description of offline object extraction can be found in \cite{Stumper18LabelExtraction}.

%% file: doc/sec_loss_function.tex
\mysection{Training}{training}
\mysubsection{Loss Function}{loss_function}
For the training of our deep neural network, we combine adapted versions of the loss functions from \cite{Hoermann17Prediction} and \cite{Hoermann18Detection} to incorporate the segmentation of static and dynamic regions into the object detection process introduced in \cite{Hoermann18Detection}. The total loss is given by
\begin{equation}
	L = L_s + L_d,
\end{equation}
with static loss function $L_s$ and dynamic loss $L_d$. For the distinction between static and dynamic grid cells, the desired output is the probability of each cell being static or not. This task can be classified as a regression problem, thus we employ the basic Euclidean Loss (or L$2$-Loss) as the static loss function
\begin{equation}
	L_s = \frac{\lambda_s}{2} \sum_{\GMcell}\left( \Netout_s(\GMcell) - \Label_s(\GMcell) \right)^2,
\end{equation}
where $\lambda_s$ denotes the static weighting factor. To tackle the high imbalance between dynamic and static grid cells, which is described in \cite{Hoermann17Prediction} and \cite{Hoermann18Detection}, the spatial balancing loss function 
\begin{equation}\label{eq:spatial_balancing_loss}
	L_d =  L_{d}^{(IoU)} + L_{d}^{(\ANCdWrel)} + L_{d}^{(\ANCdLrel)} + L_{d}^{(\ANCdOriRel)}
\end{equation}
is used \cite{Hoermann18Detection}. Each term on the right hand side in \eqref{eq:spatial_balancing_loss} is given by
\begin{equation}
\begin{split}
	L_{d}^{(\cdot)} = \frac{\lambda_d^{(\cdot)}}{2} \sum_\GMcell \sum_\Anchor (1 + \GainParam \cdot \LabelIoUMax(\GMcell)^{\FocusingParam_d}) \\
	\left( \Netout_{d}^{(\cdot)} (\GMcell,\Anchor) - \Label_{d}^{(\cdot)} (\GMcell,\Anchor) \right)^2,
\end{split}
\end{equation}
where $\Label_{d}^{(\cdot)} \in \{\LabelIoU_{d}, \LabelDW_{d}, \LabelDL_{d}, \LabelDOri_{d}\}$ and $\lambda_d^{(\cdot)}$ denotes the weighting factor for each dynamic loss to adjust the influence of each output. To adjust the weighting of the cells, $\LabelIoUMax(\GMcell)$ is used as a spatial map, where $\LabelIoUMax(\GMcell) = 0$ for all background cells and $0 < \LabelIoUMax(\GMcell) \leq 1$ for all cells that are occupied. The foreground gain $\GainParam$ reduces the aforementioned imbalance between background and object cells. Background cells are weighted by $1$, since $\LabelIoUMax(\GMcell) = 0$ and dynamic cells are weighted with a maximum of $1 + \GainParam$. The focus parameter $\FocusingParam_d$ adjusts the weighting of the cells within the bounds of an object, e.g. $0 < \LabelIoUMax(\GMcell) \leq 1$. A more in-depth explanation of all parameters can be found in \cite{Hoermann18Detection}.
Furthermore, we add a regularization term with $\lambda_R = 10^{-7}$ and dropout layers with a dropout rate of $r_d = 0.1$ to avoid over-fitting, which was particular prevalent in the detection of small static regions as dynamic objects. For back-propagation and optimization the ADAM solver \cite{ADAM} was used due to its automatic learning rate update. The exponential decay rates for the ADAM algorithm are set to $\beta_1 = 0.9$ and $\beta_2 = 0.999$ and the starting learning rate is chosen to $l = 0.0001$.
\mysubsection{Dataset}{dataset}
We collected about $2\,\si{\hour}$ of data during multiple days in an urban shared space with pedestrians, other motor vehicles and bikes, while standing in the middle of the road on a parking lane. The labels were created with the automatic label generation algorithm introduced in \cite{Hoermann18Detection}. The sequences contain a total of $78011$ images of which $68750$ were used for training, $8061$ for testing and $1200$ for the evaluation. The \DOGMA, which is used as the input, has spatial dimensions of $901 \times 901$ with a cell width of $0.15\,\si{\meter}$.

We showed in \cite{Hoermann18Detection}, that on average, the ratio of occupied to free cells in our dataset is about $1 : 31$, the ratio of occupied to not occupied is about $1 : 65$ and the ratio of dynamic foreground cells to total occupied background cells is $1 : 400$. Furthermore, we also discovered an extreme imbalance between labels with $\LabelIoU = 0$ and $\LabelIoU > 0$ and even a high imbalance between $0 < \LabelIoU < 0.5$ and labels with $\LabelIoU \geq 0.5$. To reduce these imbalances, we chose the balanced loss functions and set the foreground gain to $\GainParam = 400$, which is exactly the ratio of foreground to background cells, as described earlier. To counteract the differences within dynamic cells ($0 < \LabelIoU < 1$), the focus parameter is set to $\FocusingParam_d = 4$.
In comparison to the IoU label, the offset labels have equal values near the objects center and therefore, we conclude $\FocusingParam^{(\ANCdWrel)}_d = \FocusingParam^{(\ANCdLrel)}_d = \FocusingParam^{(\ANCdOriRel)}_d = 1$. Additionally, we chose the remaining weighting parameters as follows: $\lambda_d^{(IoU)} = 1, \lambda_d^{(\ANCdWrel)} = 0.01, \lambda_d^{(\ANCdLrel)} = 0.05, \lambda_d^{(\ANCdOriRel)} = 0.25$ and $\lambda_d^{(s)} = 0.5$.
 
The training process was stopped after about $1.5$ epochs with a total of $120000$ iterations and a batch size of $1$.

%% file: doc/sec_results.tex
\mysection{Results and Evaluation}{results}
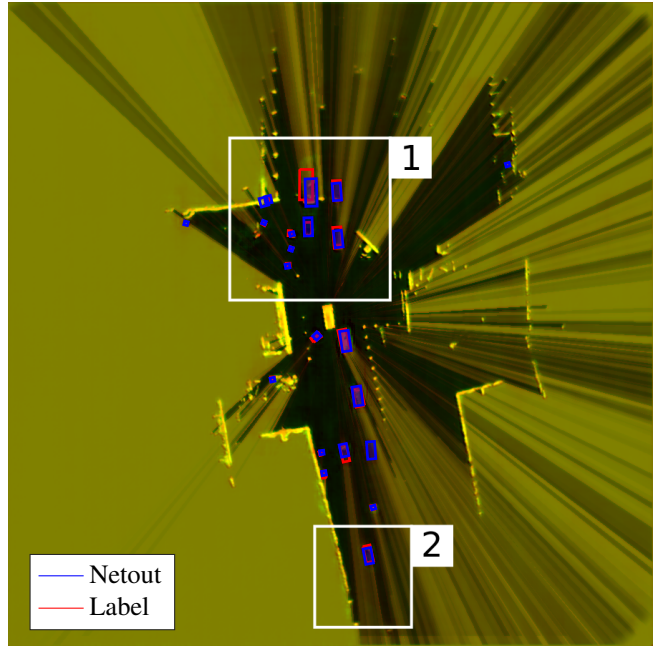
\begin{figure}[!tbp]
	\centering
	\input{img/sequence_overview.tikz}
	\caption{Overview of the scene with static and dynamic segmentation of cells depicted in yellow and black. Blue bounding boxes represent detected dynamic objects, while red boxes are the ground-truth labels. The ego vehicle is the yellow rectangle in the middle of the scene facing upwards.}
	\label{fig:scene_overview}
\end{figure}

For the evaluation, we use a dataset that was recorded on different days in the same shared space downtown environment as the training dataset. The scene is illustrated in Fig. \ref{fig:scene_overview}. Again, pedestrians, other motor vehicles and bikes were present during the time of the recording. Due to the rather random behavior of all road users, e.g. crossing the road at arbitrary locations or disappearing behind walls, our neural network has to deal with occlusion and predict and track the movement of objects by exploiting the sequential input data. The dataset was recorded by four Velodyne VLP-16 and one IBEO LUX laser scanner. We obtain a $360^{\circ}$ perception of the environment with our sensor setup and captured data in the range of $200\,\si{\meter}$ (with opening angle of $100^{\circ}$ and four layers) with the IBEO LUX laser scanner. The \mbox{Velodyne VLP-16} provide 16 layers of laser measurements with a range of $100\,\si{\meter}$. We use about $2\,\si{\minute}$ from two sequences for the evaluation. At the end of this section we also show the results of the same network, that was trained with only the Dempster-Shafer masses for free space $\MassFree$ and occupancy $\MassOcc$. Our current input is generated by Bayesian sensor fusion and provides not only the masses but also the velocities and the corresponding variances for each cell.

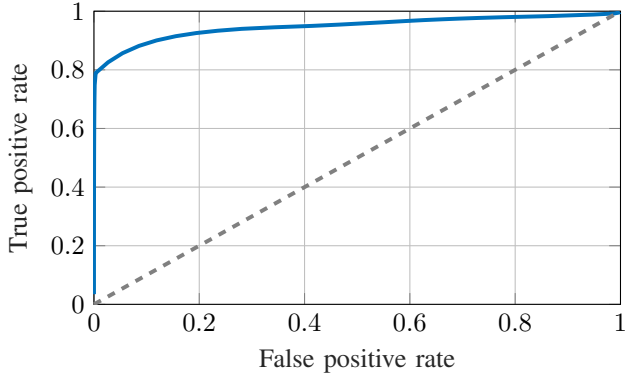
\begin{figure}[!tbp]
	\centering
	\input{img/roc.tikz}
	\caption{Receiver operating characteristic curve (ROC) for segmentation between static and dynamic cells. The gray dashed line is the line of no-discrimination, which is equal to random guessing. The greater the distance between this line and the ROC curve, the better the classifier.}
	\label{fig:roc_curve}
\end{figure}

To asses the performance of our classifier for the decision between dynamic and static cells, the receiver operator characteristic (ROC) curve is plotted in Fig. \ref{fig:roc_curve} by varying the threshold for the IoU between $0.01$ and $1$ and examining the true positive rate and the false positive rate. The area under curve (AUC) score for the classifier is $0.9463$, with values between $0.9$ and $1$ usually denoting an excellent classification result. However, this segmentation is an easy task for most artificial neural networks and many cells have an occupancy probability of $P_\Occupied = 0.5$, which means that their state is unknown. Nevertheless, the classifier yields satisfactory results in significant regions, which can be seen in Figure \ref{fig:scene_overview}, where yellow cells show the static regions, black cells represent free space and red cells are dynamic objects.

\begin{table}[!tbp]
	\renewcommand{\arraystretch}{1.3}
	\caption{Bounding box error (RMSE) and average precision (AP)}
	\label{tab:rmse_ap}
	\centering
	\begin{tabularx}{\columnwidth}{lccccc}
		\toprule
		& \textbf{width} & \textbf{length} & \textbf{position}  &  \textbf{orientation} & \textbf{AP} \\
		\midrule 
		\textbf{old} & $0.20\,\si{\meter}$ & $0.70\,\si{\meter}$  & $0.48\,\si{\meter}$ & $8.9^{\circ}$ & $0.75$\\
		\textbf{new} & \colorbox{green}{$0.18\,\si{\meter}$} & \colorbox{green}{$0.57\,\si{\meter}$}  & $0.43\,\si{\meter}$ & $9.1^{\circ}$ & $0.78$ \\
		\bottomrule
	\end{tabularx}
\end{table}

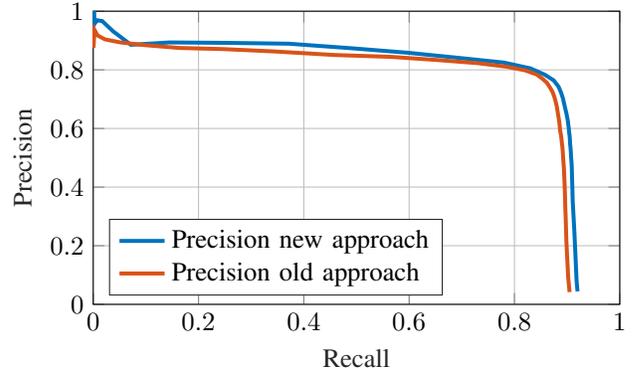
\begin{figure}[!t]
	\centering
	\input{img/precision_recall.tikz}
	\caption{Comparison of object detection precision recall curves between our previous approach and the new contribution with a CNN and a RNN.}
	\label{fig:precision_recall}
\end{figure}

\begin{figure*}[!ht]
	\centering
	\input{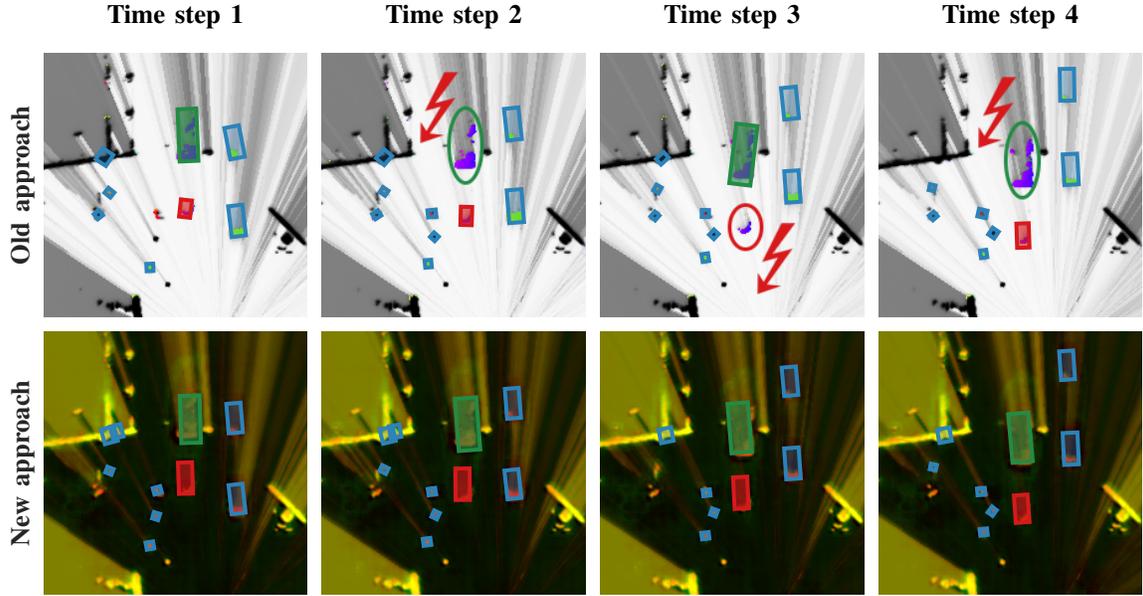}
	\caption{Excerpt of the grid map in front of the ego-vehicle. The top sequence shows the result of our old approach, while the bottom sequence was created with our new approach. The green and red objects, that are detected in the first frame in both approaches, are of interest. In time step $2$ of our old method, the green object is occluded by the red object and is not detected, which is visualized by the green circle and the red lightning bold symbol. In time step $3$ the green object is detected again with a different size and orientation estimate. Furthermore, the red detection is lost, which is again visualized by the red circle and the lightning bold symbol. In the last frame (time step $4$), the green object is occluded and misdetected again. With our new approach, both the green and red objects are continuously detected in the bottom sequence in all frames and have a more consistent size estimate.  }
	\label{fig:seq_1}
\end{figure*}

\begin{figure*}[!ht]
	\centering
	\input{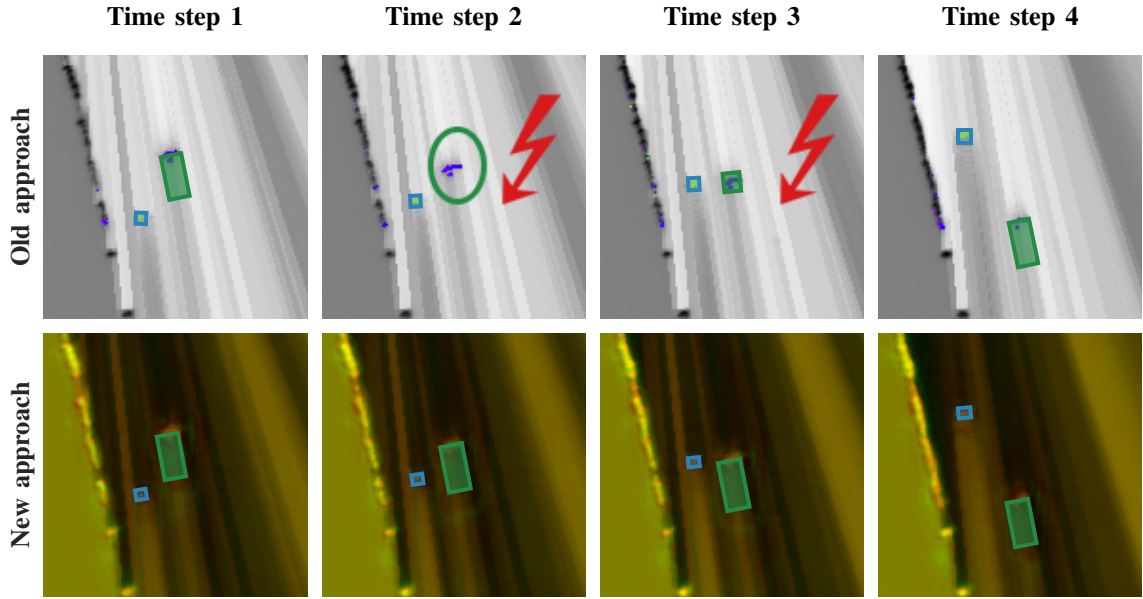}
	\caption{Far distant perception of the environment (about $50\,\si{\meter}$ behind the ego-vehicle). The top sequence shows our old approach and the bottom sequence depicts the results of our new method.  The green object, that is detected with both approaches in the first time step is of interest. With our old network, the detection is lost in time step $2$ due to very few measurements and is picked up again in time step $3$ with an obvious wrong size estimation. This is visualized by the red lightning bold symbol. Again, with our new approach, we get a continuous detection of the green object during the whole sequence shown in the bottom row. Additionally, the consistency of the size estimate is improved significantly.}
	\label{fig:seq_2}
\end{figure*}

To evaluate our results against our previous approach, the precision recall curve for the network with only a CNN and the new network combining a CNN and RNN is depicted in Fig. \ref{fig:precision_recall}. The curve shows the object detection performance and was created by varying the threshold for the IoU between $0.01$ and $1$ and calculating the precision and the true positive rate (recall). We gain a slight but consistent improvement with our proposed network structure compared to our old approach for every IoU threshold with an average precision of $0.78$ in contrast to $0.75$. Furthermore, the root mean square errors for the position, width, length and orientation of the bounding boxes of our old and our new network are listed in Table \ref{tab:rmse_ap}. The position, width and length estimates are improved due to the temporal filtering of the LSTM cell. To demonstrate these effects, two exemplary scenes are illustrated in Fig. \ref{fig:seq_1} and \ref{fig:seq_2}, that correspond to the regions $1$ and $2$ in Fig. \ref{fig:scene_overview} respectively. Fig. \ref{fig:seq_1} shows the results for an object that gets occluded by another one. The top row depicts the previous approach, while the bottom row shows the result of our new proposed approach. With our recurrent neural network, the green and red objects are detected in all four time steps, even when the object in the front covers the one in the back. With only the CNN (top row), the network fails to detect the rear object when it gets occluded in time step $2$, even though it detected both objects in the first frame. This misbehavior may be traced back to not having a memory of previous time steps, where our new approach remembers a detected object and tracks it over time. Another advantage of the memory is the consistency of the size estimates of the bounding boxes. When our old network loses a detection in the second frame and detects it again in the third frame, it has no knowledge about its past state and both the dimension and orientation estimates get worse. Again, the old approach fails to detect the green object in time step $4$. This behavior can lead to accidents in the real world, when dynamic obstacles like cars or pedestrians get overlooked by the object detection algorithm. Additionally, the varying size estimation can also be observed in the second scene in Fig. \ref{fig:seq_2}, where a far distant perception is depicted. The RNN tracks the moving object and keeps its size consistent over time, while the network without memory loses the detection in the second frame. Furthermore, the network underestimates the dimensions before getting the correct size again in the last time step. A wrong size estimation is particular hazardous for, e.g., behavior or trajectory planning and prediction, when the system assumes free space but in reality the object dimensions were wrongfully estimated. 

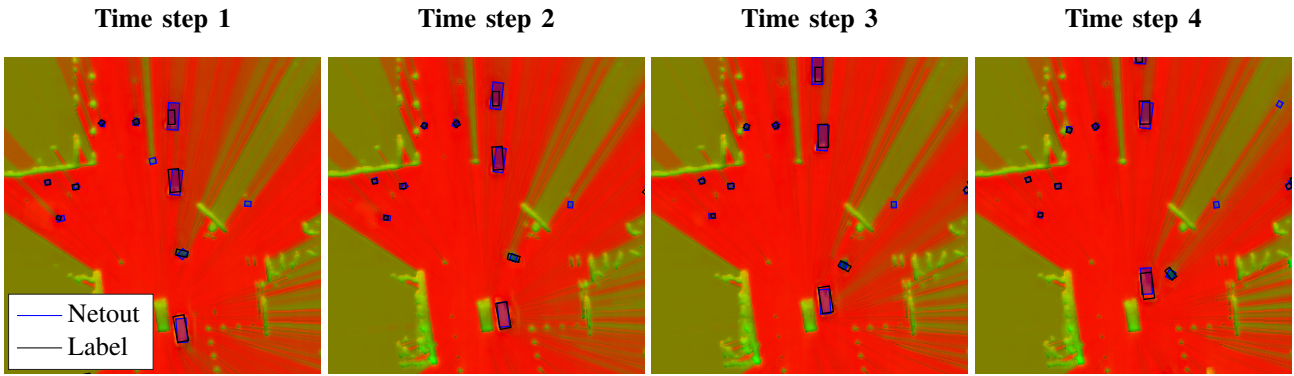
\begin{figure*}[!ht]
	\centering
	\input{img/sequence3/s3.tikz}
	\caption{Excerpt of the results without velocities at input. Objects detected by the network are depicted as blue bounding boxes and ground-truth labels are shown as black rectangles. The results show satisfactory and unvarying object hypotheses, that are consistent with the ground-truth labels. The estimation of static and dynamic cells fails. Most of the cells are estimated as dynamic (red regions), which obviously is a false classification.}
	\label{fig:seq_3}
\end{figure*}

In addition to our proposed method, we also conducted a preliminary experiment without the velocity channels obtained from the particle filter as input. Only the masses for occupancy $(\MassOcc)$ and free space $(\MassFree)$ were used to train the network. Note that for the calculation of the masses the velocities were used, however, they are not fed explicitly to the network anymore. An exemplary scene from this experiment is depicted in Fig. \ref{fig:seq_3}. The results confirm that relevant features can be extracted from the Dempster-Shafer masses alone, so that objects are detected and parameterized. The object hypotheses are promising and show satisfactory and consistent bounding boxes in comparison to the ground-truth labels. However, one limitation so far, is the misclassification of dynamic and static cells without information about the dynamics of the scene.


%% file: img/sequence_overview.tikz
%
%
\begin{tikzpicture}

\begin{axis}[%
width=9.5cm,
height=9.5cm,
at={(0,0)},
scale only axis,
axis on top,
unbounded coords=jump,
xmin=0,
xmax=1000,
tick align=outside,
y dir=reverse,
ymin=0,
ymax=1000,
axis line style={draw=none},
ticks=none,
axis x line*=bottom,
axis y line*=left,
legend style={at={(0.03,0.12)}, anchor=south west, legend cell align=left, align=left, draw=white!15!black}
]
\addplot [forget plot] graphics [xmin=0.5, xmax=901.5, ymin=0.5, ymax=901.5] {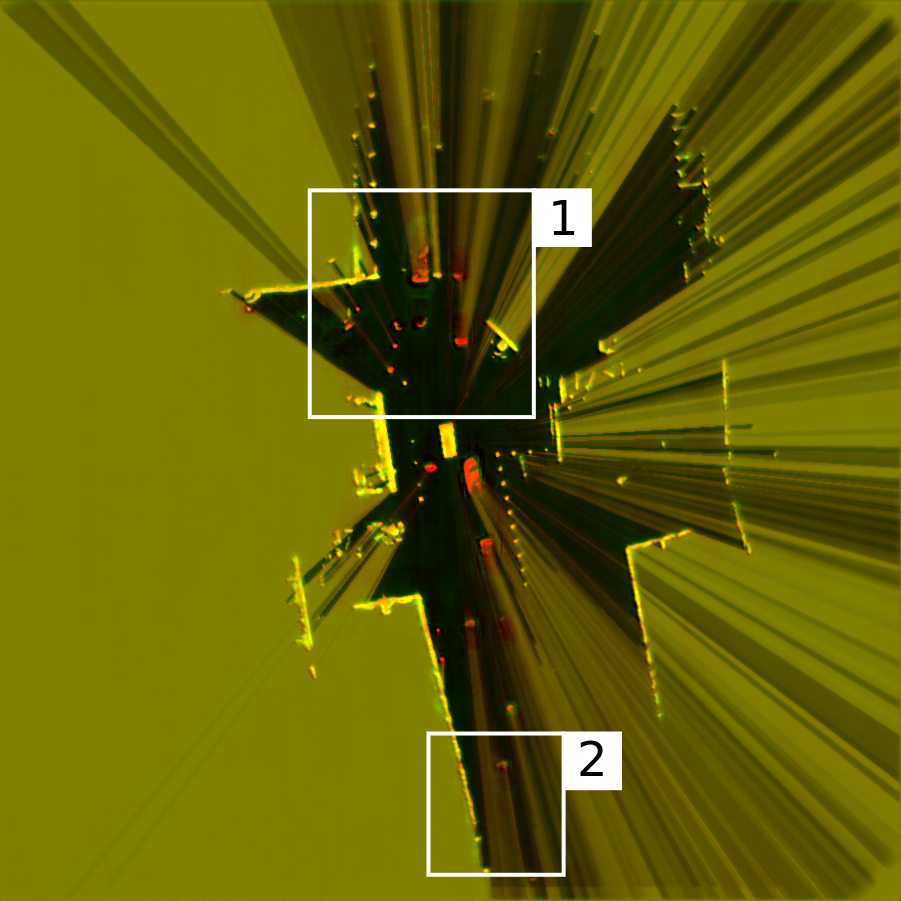};

\addplot [color=red, line width=1.0pt, forget plot]
table[row sep=crcr]{%
	361.698516845703	309.176849365234\\
	356.831451416016	306.395660400391\\
	354.301483154297	310.823150634766\\
	359.168548583984	313.604339599609\\
	361.698516845703	309.176849365234\\
};
\addplot [color=red, line width=1.0pt, forget plot]
table[row sep=crcr]{%
	358.568969726562	278.9111328125\\
	362.766296386719	282.517700195312\\
	367.431030273438	277.0888671875\\
	363.233703613281	273.482299804688\\
	358.568969726562	278.9111328125\\
};
\addplot [color=red, line width=1.0pt, forget plot]
table[row sep=crcr]{%
	372.736602783203	530.720581054688\\
	372.514343261719	525.07275390625\\
	367.263397216797	525.279418945312\\
	367.485656738281	530.92724609375\\
	372.736602783203	530.720581054688\\
};
\addplot [color=red, line width=1.0pt, forget plot]
table[row sep=crcr]{%
	394.767364501953	372.029449462891\\
	394.585235595703	365.813751220703\\
	389.232635498047	365.970550537109\\
	389.414764404297	372.186248779297\\
	394.767364501953	372.029449462891\\
};
\addplot [color=red, line width=1.0pt, forget plot]
table[row sep=crcr]{%
	398.317840576172	320.523529052734\\
	392.068450927734	320.192138671875\\
	391.682159423828	327.476470947266\\
	397.931549072266	327.807861328125\\
	398.317840576172	320.523529052734\\
};
\addplot [color=red, line width=1.0pt, forget plot]
table[row sep=crcr]{%
	396.628936767578	349.226531982422\\
	399.052886962891	344.781005859375\\
	395.371063232422	342.773468017578\\
	392.947113037109	347.218994140625\\
	396.628936767578	349.226531982422\\
};
\addplot [color=red, line width=1.0pt, forget plot]
table[row sep=crcr]{%
	408.028411865234	234.567749023438\\
	407.678283691406	277.282318115234\\
	425.971588134766	277.432250976562\\
	426.321716308594	234.717697143555\\
	408.028411865234	234.567749023438\\
};
\addplot [color=red, line width=1.0pt, forget plot]
table[row sep=crcr]{%
	414.877136230469	308.929290771484\\
	415.324279785156	327.30908203125\\
	425.122863769531	327.070709228516\\
	424.675720214844	308.69091796875\\
	414.877136230469	308.929290771484\\
};
\addplot [color=red, line width=1.0pt, forget plot]
table[row sep=crcr]{%
	427.615264892578	476.791717529297\\
	438.511993408203	468.925872802734\\
	434.384735107422	463.208282470703\\
	423.488006591797	471.074127197266\\
	427.615264892578	476.791717529297\\
};
\addplot [color=red, line width=1.0pt, forget plot]
table[row sep=crcr]{%
	441.216766357422	633.797180175781\\
	439.643829345703	628.702209472656\\
	434.783233642578	630.202819824219\\
	436.356170654297	635.297790527344\\
	441.216766357422	633.797180175781\\
};
\addplot [color=red, line width=1.0pt, forget plot]
table[row sep=crcr]{%
	445.494873046875	666.281066894531\\
	443.862335205078	656.796752929688\\
	438.505126953125	657.718933105469\\
	440.137664794922	667.203247070312\\
	445.494873046875	666.281066894531\\
};
\addplot [color=red, line width=1.0pt, forget plot]
table[row sep=crcr]{%
	466.186859130859	277.484649658203\\
	464.976135253906	250.023208618164\\
	453.813140869141	250.515365600586\\
	455.023864746094	277.976776123047\\
	466.186859130859	277.484649658203\\
};
\addplot [color=red, line width=1.0pt, forget plot]
table[row sep=crcr]{%
	467.835693359375	344.160766601562\\
	465.68701171875	314.990478515625\\
	454.164306640625	315.839233398438\\
	456.31298828125	345.009521484375\\
	467.835693359375	344.160766601562\\
};
\addplot [color=red, line width=1.0pt, forget plot]
table[row sep=crcr]{%
	478.732696533203	487.721984863281\\
	472.483764648438	457.925689697266\\
	461.267303466797	460.278015136719\\
	467.516235351562	490.074310302734\\
	478.732696533203	487.721984863281\\
};
\addplot [color=red, line width=1.0pt, forget plot]
table[row sep=crcr]{%
	463.801788330078	621.465515136719\\
	468.870330810547	644.579895019531\\
	478.198211669922	642.534484863281\\
	473.129669189453	619.420104980469\\
	463.801788330078	621.465515136719\\
};
\addplot [color=red, line width=1.0pt, forget plot]
table[row sep=crcr]{%
	498.916198730469	564.974365234375\\
	493.122406005859	538.40087890625\\
	481.083801269531	541.025634765625\\
	486.877593994141	567.59912109375\\
	498.916198730469	564.974365234375\\
};
\addplot [color=red, line width=1.0pt, forget plot]
table[row sep=crcr]{%
	496.336151123047	762.839294433594\\
	501.595520019531	785.497680664062\\
	511.663848876953	783.160705566406\\
	506.404479980469	760.502319335938\\
	496.336151123047	762.839294433594\\
};
\addplot [color=red, line width=1.0pt, forget plot]
table[row sep=crcr]{%
	514.603515625	710.410827636719\\
	513.886108398438	704.764587402344\\
	507.396514892578	705.589172363281\\
	508.113922119141	711.235412597656\\
	514.603515625	710.410827636719\\
};

\addplot[area legend, line width=1.0pt, draw=blue, fill=blue, fill opacity=0.329536688327789, forget plot]
table[row sep=crcr] {%
x	y\\
245.734191894531	311.924194335938\\
250.882568359375	313.289978027344\\
252.265808105469	308.075805664062\\
247.117431640625	306.710021972656\\
245.734191894531	311.924194335938\\
}--cycle;

\addplot[area legend, line width=1.0pt, draw=blue, fill=blue, fill opacity=0.262440097332001, forget plot]
table[row sep=crcr] {%
x	y\\
351.129364013672	275.48876953125\\
354.353271484375	287.365997314453\\
364.870635986328	284.51123046875\\
361.646728515625	272.634002685547\\
351.129364013672	275.48876953125\\
}--cycle;

\addplot[area legend, line width=1.0pt, draw=blue, fill=blue, fill opacity=0.40139878988266, forget plot]
table[row sep=crcr] {%
x	y\\
361.451873779297	307.644287109375\\
356.371917724609	305.667938232422\\
354.548126220703	310.355712890625\\
359.628082275391	312.332061767578\\
361.451873779297	307.644287109375\\
}--cycle;

\addplot[area legend, line width=1.0pt, draw=blue, fill=blue, fill opacity=0.295478057861328, forget plot]
table[row sep=crcr] {%
x	y\\
358.527618408203	273.010711669922\\
361.055908203125	283.098663330078\\
369.472381591797	280.989288330078\\
366.944091796875	270.901336669922\\
358.527618408203	273.010711669922\\
}--cycle;

\addplot[area legend, line width=1.0pt, draw=blue, fill=blue, fill opacity=0.306590569019318, forget plot]
table[row sep=crcr] {%
x	y\\
373.601379394531	531.173950195312\\
372.380157470703	525.531433105469\\
366.398620605469	526.826049804688\\
367.619842529297	532.468566894531\\
373.601379394531	531.173950195312\\
}--cycle;

\addplot[area legend, line width=1.0pt, draw=blue, fill=blue, fill opacity=0.352172207832336, forget plot]
table[row sep=crcr] {%
x	y\\
394.552307128906	372.372222900391\\
393.561645507812	366.581787109375\\
387.447692871094	367.627777099609\\
388.438354492188	373.418212890625\\
394.552307128906	372.372222900391\\
}--cycle;

\addplot[area legend, line width=1.0pt, draw=blue, fill=blue, fill opacity=0.321231436729431, forget plot]
table[row sep=crcr] {%
x	y\\
397.648193359375	349.409362792969\\
399.304107666016	344.149810791016\\
394.351806640625	342.590637207031\\
392.695892333984	347.850189208984\\
397.648193359375	349.409362792969\\
}--cycle;

\addplot[area legend, line width=1.0pt, draw=blue, fill=blue, fill opacity=0.306089890003204, forget plot]
table[row sep=crcr] {%
x	y\\
399.328125	321.555328369141\\
393.219970703125	323.268707275391\\
394.671875	328.444671630859\\
400.780029296875	326.731292724609\\
399.328125	321.555328369141\\
}--cycle;

\addplot[area legend, line width=1.0pt, draw=blue, fill=blue, fill opacity=0.336247456073761, forget plot]
table[row sep=crcr] {%
x	y\\
413.820068359375	302.434967041016\\
414.628356933594	327.931243896484\\
426.179931640625	327.565032958984\\
425.371643066406	302.068756103516\\
413.820068359375	302.434967041016\\
}--cycle;

\addplot[area legend, line width=1.0pt, draw=blue, fill=blue, fill opacity=0.322683584690094, forget plot]
table[row sep=crcr] {%
x	y\\
415.4013671875	247.964736938477\\
416.580871582031	286.525207519531\\
432.5986328125	286.035278320312\\
431.419128417969	247.47477722168\\
415.4013671875	247.964736938477\\
}--cycle;

\addplot[area legend, line width=1.0pt, draw=blue, fill=blue, fill opacity=0.413889062404633, forget plot]
table[row sep=crcr] {%
x	y\\
431.888702392578	474.1181640625\\
438.118286132812	467.896148681641\\
432.111297607422	461.8818359375\\
425.881713867188	468.103851318359\\
431.888671875	474.1181640625\\
}--cycle;

\addplot[area legend, line width=1.0pt, draw=blue, fill=blue, fill opacity=0.369425249099731, forget plot]
table[row sep=crcr] {%
x	y\\
442.317230224609	633.456481933594\\
441.613891601562	627.805358886719\\
435.682769775391	628.543518066406\\
436.386108398438	634.194641113281\\
442.317230224609	633.456481933594\\
}--cycle;

\addplot[area legend, line width=1.0pt, draw=blue, fill=blue, fill opacity=0.4062828540802, forget plot]
table[row sep=crcr] {%
x	y\\
445.560974121094	662.484985351562\\
444.527526855469	656.469055175781\\
438.439025878906	657.515014648438\\
439.472473144531	663.530944824219\\
445.560974121094	662.484985351562\\
}--cycle;

\addplot[area legend, line width=1.0pt, draw=blue, fill=blue, fill opacity=0.352258932590485, forget plot]
table[row sep=crcr] {%
x	y\\
466.776092529297	277.546508789062\\
464.811889648438	253.506683349609\\
453.223907470703	254.453491210938\\
455.188110351562	278.493316650391\\
466.776092529297	277.546508789062\\
}--cycle;

\addplot[area legend, line width=1.0pt, draw=blue, fill=blue, fill opacity=0.356217098236084, forget plot]
table[row sep=crcr] {%
x	y\\
468.962799072266	343.173034667969\\
466.346466064453	319.573211669922\\
455.037200927734	320.826965332031\\
457.653533935547	344.426788330078\\
468.962799072266	343.173034667969\\
}--cycle;

\addplot[area legend, line width=1.0pt, draw=blue, fill=blue, fill opacity=0.281816673278809, forget plot]
table[row sep=crcr] {%
x	y\\
463.124694824219	620.445922851562\\
466.620727539062	637.639221191406\\
476.875305175781	635.554077148438\\
473.379272460938	618.360778808594\\
463.124694824219	620.445922851562\\
}--cycle;

\addplot[area legend, line width=1.0pt, draw=blue, fill=blue, fill opacity=0.393314969539642, forget plot]
table[row sep=crcr] {%
x	y\\
480.237243652344	488.33447265625\\
475.986206054688	459.841796875\\
463.762756347656	461.66552734375\\
468.013793945312	490.158203125\\
480.237243652344	488.33447265625\\
}--cycle;

\addplot[area legend, line width=1.0pt, draw=blue, fill=blue, fill opacity=0.353564393520355, forget plot]
table[row sep=crcr] {%
x	y\\
496.981292724609	564.498657226562\\
493.199401855469	537.777404785156\\
481.018707275391	539.501342773438\\
484.800598144531	566.222595214844\\
496.981292724609	564.498657226562\\
}--cycle;

\addplot[area legend, line width=1.0pt, draw=blue, fill=blue, fill opacity=0.271580529212952, forget plot]
table[row sep=crcr] {%
x	y\\
496.590118408203	766.185913085938\\
500.616912841797	787.82275390625\\
511.409881591797	785.814086914062\\
507.383087158203	764.17724609375\\
496.590118408203	766.185913085938\\
}--cycle;

\addplot[area legend, line width=1.0pt, draw=blue, fill=blue, fill opacity=0.262126338481903, forget plot]
table[row sep=crcr] {%
x	y\\
514.60595703125	639.685119628906\\
512.743591308594	615.442993164062\\
501.394073486328	616.314880371094\\
503.256408691406	640.557006835938\\
514.60595703125	639.685119628906\\
}--cycle;

\addplot[area legend, line width=1.0pt, draw=blue, fill=blue, fill opacity=0.319666874408722, forget plot]
table[row sep=crcr] {%
x	y\\
514.693786621094	709.678039550781\\
513.131774902344	704.548095703125\\
507.306182861328	706.321960449219\\
508.868194580078	711.451904296875\\
514.693786621094	709.678039550781\\
}--cycle;

\addplot[area legend, line width=1.0pt, draw=blue, fill=blue, fill opacity=0.366791796684265, forget plot]
table[row sep=crcr] {%
x	y\\
702.370544433594	229.801559448242\\
701.057556152344	224.779373168945\\
695.629455566406	226.198440551758\\
696.942443847656	231.220626831055\\
702.370544433594	229.801559448242\\
}--cycle;

\addplot [color=blue, line width=1.5pt, forget plot]
  table[row sep=crcr]{%
0	0\\
};
\addplot [color=red, line width=1.0pt, forget plot]
  table[row sep=crcr]{%
0	0\\
};
\addplot [color=blue]
  table[row sep=crcr]{%
0	0\\
};
\addlegendentry{Netout}

\addplot [color=red]
  table[row sep=crcr]{%
0	0\\
};
\addlegendentry{Label}

\end{axis}
\end{tikzpicture}%

%% file: img/roc.tikz
%
%
\definecolor{mycolor1}{rgb}{0.00000,0.44700,0.74100}%
\begin{tikzpicture}

\begin{axis}[%
width=7cm,
height=3.9cm,
at={(0.772in,0.516in)},
scale only axis,
xmin=0,
xmax=1,
xlabel style={font=\color{white!15!black}},
xlabel={False positive rate},
ymin=0,
ymax=1,
ylabel style={font=\color{white!15!black}},
ylabel={True positive rate},
axis background/.style={fill=white},
xmajorgrids,
ymajorgrids,
legend style={legend cell align=left, align=left, draw=white!15!black}
]
\addplot [color=mycolor1,  line width=1.5pt]
  table[row sep=crcr]{%
0.99904884792511	0.997610318188055\\
0.998325554149206	0.996695836170469\\
0.996533548643885	0.995225352374625\\
0.988264294984824	0.992320970438485\\
0.948427749010295	0.988605456178409\\
0.863641240867386	0.98330886850395\\
0.776872077363577	0.979518543297918\\
0.730210383241753	0.977073903877041\\
0.694879236444954	0.9748481351108\\
0.661375932745441	0.972477523749965\\
0.625936768985109	0.969680146843474\\
0.590308266901216	0.966351373453053\\
0.552218052810186	0.962488527644866\\
0.509864222030873	0.958460428388251\\
0.465877427897934	0.954304057885146\\
0.414142231098216	0.949939086725004\\
0.343068800942729	0.945171117460748\\
0.28038533947128	0.93980845111586\\
0.235179005639212	0.933453870987139\\
0.194222678702763	0.925481620360641\\
0.15537324232759	0.914763256258647\\
0.11887816050644	0.90039014766018\\
0.0854162765314727	0.881596853643755\\
0.0543043330630469	0.856835359449036\\
0.0268733738007811	0.825833504548068\\
0.00326743956971104	0.788979375153549\\
0.00141574794546391	0.747979845970009\\
0.00104208065125679	0.706802601340189\\
0.000796354936253811	0.665986426246931\\
0.000621796335041907	0.626220858880699\\
0.000490842973798467	0.587380083068169\\
0.000387163778636207	0.549766296497355\\
0.000307553998609304	0.512607028086675\\
0.000245126460537928	0.475720830045566\\
0.000196768442924376	0.439479729272141\\
0.000158229939055463	0.404047646320092\\
0.00012755758933296	0.369057124639924\\
0.000105585527711022	0.334576978561421\\
8.91771739552595e-05	0.301023378281724\\
7.59817518716737e-05	0.268645750509822\\
6.63038967531174e-05	0.237020571770269\\
5.83418334162289e-05	0.206657016857512\\
5.1965502941036e-05	0.178239672220036\\
4.65979077082237e-05	0.152191952313699\\
4.20005630388026e-05	0.128283573110023\\
3.79416864304609e-05	0.106699078017976\\
3.43903136671842e-05	0.087669718282137\\
3.13196739174641e-05	0.0711178447830317\\
2.85493843073564e-05	0.0567748020997211\\
2.60106402922632e-05	0.0446936999822954\\
2.36684666223825e-05	0.034762540054528\\
};

\addplot [color=gray, dashed,  line width=1.5pt]
  table[row sep=crcr]{%
0	0\\
1	1\\
};

\end{axis}
\end{tikzpicture}%

%% file: img/precision_recall.tikz
%
%
\definecolor{mycolor1}{rgb}{0.00000,0.44700,0.74100}%
\definecolor{mycolor2}{rgb}{0.85000,0.32500,0.09800}%
\begin{tikzpicture}

\begin{axis}[%
width=7cm,
height=3.9cm,
at={(0.758in,0.481in)},
scale only axis,
unbounded coords=jump,
xmin=0,
xmax=1,
xlabel style={font=\color{white!15!black}},
xlabel={Recall},
ymin=0,
ymax=1,
ylabel style={font=\color{white!15!black}},
ylabel={Precision},
axis background/.style={fill=white},
xmajorgrids,
ymajorgrids,
legend style={at={(0.03,0.03)}, anchor=south west, legend cell align=left, align=left, draw=white!15!black}
]
\addplot [color=mycolor1, line width=1.5pt]
  table[row sep=crcr]{%
0.919804442697238	0.044150754185218\\
0.917994268690903	0.0800615548357482\\
0.913824454206317	0.247003945229422\\
0.910631910019581	0.351819774628235\\
0.909806793930221	0.408501516884366\\
0.909271375930514	0.439823828338744\\
0.90890656639885	0.46561065772243\\
0.908398730717747	0.484943879594\\
0.907483131329971	0.50303620610798\\
0.907063242839582	0.519587396937833\\
0.90651666872724	0.533798931682915\\
0.905564713201815	0.548788639823138\\
0.905345143099505	0.562211050767905\\
0.90473375597669	0.575667485954239\\
0.903750027712587	0.588384298145459\\
0.903199225635116	0.600450343837062\\
0.902233176225136	0.612984021846735\\
0.901484200777568	0.625195218732102\\
0.900211173750217	0.637148816586816\\
0.898817897368372	0.650614745181513\\
0.896840827541667	0.665189656574176\\
0.894752861071168	0.681093759403892\\
0.892343161806366	0.70094025072866\\
0.888885081611245	0.720705192961522\\
0.883881973108223	0.742222889307491\\
0.874536605166617	0.764087035770361\\
0.857842578613826	0.783225762143269\\
0.830163874744241	0.805124567068848\\
0.77989202533836	0.824845303418198\\
0.698967844817058	0.839908098733575\\
0.597847428882789	0.858263289892344\\
0.489002193125405	0.873879423132679\\
0.371724990099273	0.889161822528301\\
0.254485958900046	0.892059029177805\\
0.144363563509356	0.893480258988006\\
0.0716142620542926	0.885495741819812\\
0.0375882656576614	0.931806170198216\\
0.01695589539523	0.966581473793461\\
0.00697905367343423	0.969181340747606\\
0.00240566471445398	0.956515151515152\\
0.000485846651986276	1.07\\
};
\addlegendentry{Precision new approach}

\addplot [color=mycolor2, line width=1.5pt]
  table[row sep=crcr]{%
0.904682051838976	0.0417987427633154\\
0.90203315491474	0.09176624373917\\
0.898697959574548	0.212292461530326\\
0.897325267741356	0.294221410905742\\
0.896454501251586	0.335080800510218\\
0.896107748070923	0.366160458296294\\
0.895589810791612	0.393038167837376\\
0.895077567909257	0.416663409149428\\
0.894459853323732	0.439445730809829\\
0.8940278287257	0.461293245341145\\
0.893240375792052	0.481039231437894\\
0.892436108619743	0.499568854707261\\
0.891898495417264	0.517711098443649\\
0.891081546959537	0.536092410416849\\
0.88981979526866	0.55493470541295\\
0.88860125943088	0.574688198500107\\
0.886827945711367	0.591811883026309\\
0.885895662820524	0.609161126274197\\
0.88485019713511	0.625696381765952\\
0.883375232853357	0.641618072115621\\
0.881864852926594	0.658862018204829\\
0.880077418652112	0.677949998997329\\
0.87760340646314	0.695856080152437\\
0.875283660666831	0.710482908217394\\
0.872111828643773	0.725554540010601\\
0.866954558479017	0.741247719381984\\
0.861522861916612	0.756191007310995\\
0.853245570960115	0.769467522365955\\
0.84232856168039	0.784136798717864\\
0.819561577634467	0.798684549674095\\
0.784921097470145	0.810662369381039\\
0.730400114089526	0.82282401349975\\
0.655937171841842	0.83316248183558\\
0.564957646141378	0.844339076945138\\
0.45717500099537	0.850749522760111\\
0.344413567397458	0.862762042642314\\
0.246658083253065	0.87079454465529\\
0.1635759156187	0.874637803724105\\
0.0987963430168891	0.884137983391859\\
0.0512354251927265	0.893696131354026\\
0.0222246183340071	0.904039756967375\\
0.00796793256337079	0.917681615525833\\
0.00241437855029426	0.936634615384615\\
0.000519728483230033	0.875416666666667\\
0.000145999610667705	nan\\
};
\addlegendentry{Precision old approach}

\end{axis}
\end{tikzpicture}%

%% file: img/sequence3/s3.tikz
%
%
\begin{tikzpicture}

\definecolor{mygreen}{RGB}{0,0,0}

\definecolor{myred}{RGB}{215,25,28}
\definecolor{myblue}{RGB}{0,0,255}

\def\imagesize{4.2}

\def\offzero{0}
\def\offone{4.3}
\def\offtwo{8.6}
\def\offthree{12.9}

\begin{axis}[%
width=\imagesize cm,
height=\imagesize cm,
at={(\offzero cm, \offzero cm)},
scale only axis,
axis on top,
unbounded coords=jump,
xmin=280.134128719563,
xmax=619.234474740324,
tick align=outside,
y dir=reverse,
ymin=163.759378798304,
ymax=502.859724819065,
axis line style={draw=none},
ticks=none,
title style={font=\bfseries},
title={Time step 1},
axis x line*=bottom,
axis y line*=left,
legend style={at={(0.01,0.01)}, anchor=south west, legend cell align=left, align=left, draw=white!15!black}
]
\addplot [forget plot] graphics [xmin=0.5, xmax=901.5, ymin=0.5, ymax=901.5] {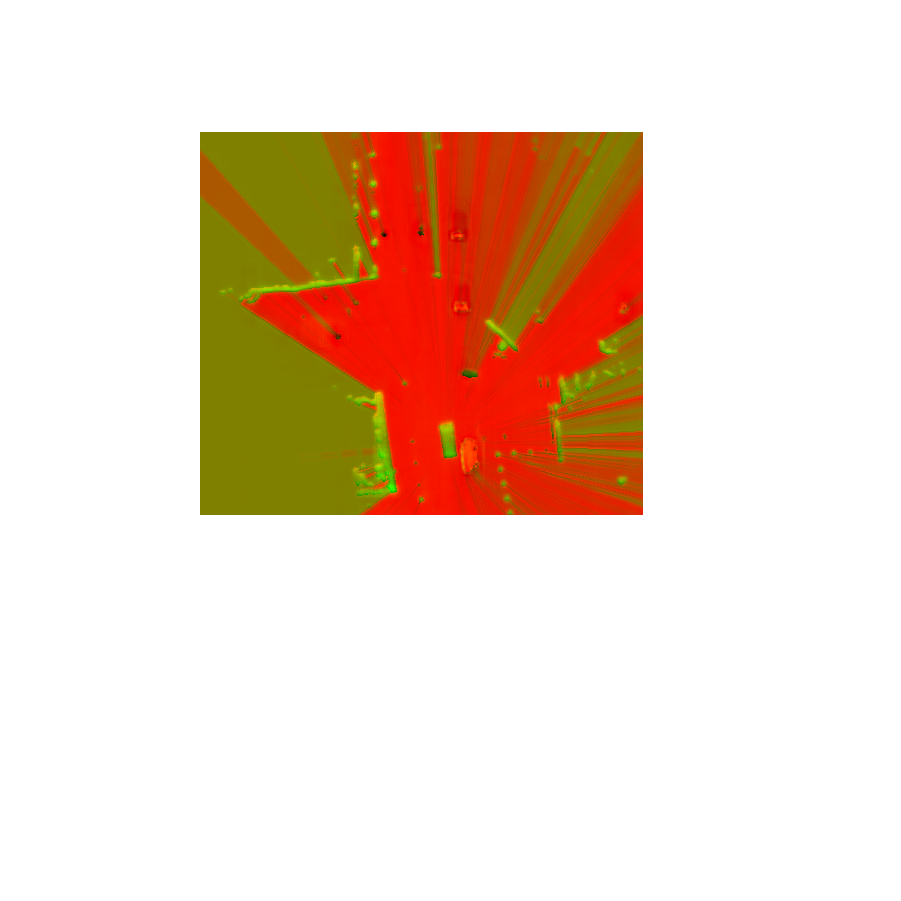};

\addplot[area legend, line width=0.5pt, draw=myblue, fill=myblue, fill opacity=0.3, forget plot]
table[row sep=crcr] {%
x	y\\
323.743957519531	301.163269042969\\
329.579925537109	299.510009765625\\
328.256042480469	294.836730957031\\
322.420074462891	296.489990234375\\
323.743957519531	301.163269042969\\
}--cycle;

\addplot[area legend, line width=0.5pt, draw=myblue, fill=myblue, fill opacity=0.3, forget plot]
table[row sep=crcr] {%
x	y\\
343.415832519531	333.311950683594\\
335.232086181641	335.407257080078\\
336.584167480469	340.688049316406\\
344.767913818359	338.592742919922\\
343.415832519531	333.311950683594\\
}--cycle;

\addplot[area legend, line width=0.5pt, draw=myblue, fill=myblue, fill opacity=0.3, forget plot]
table[row sep=crcr] {%
x	y\\
355.551239013672	305.580963134766\\
360.775787353516	302.81591796875\\
358.448760986328	298.419036865234\\
353.224212646484	301.18408203125\\
355.551239013672	305.580963134766\\
}--cycle;

\addplot[area legend, line width=0.5pt, draw=myblue, fill=myblue, fill opacity=0.3, forget plot]
table[row sep=crcr] {%
x	y\\
375.774353027344	97.6896209716797\\
377.210266113281	103.541114807129\\
382.225646972656	102.31037902832\\
380.789733886719	96.4588851928711\\
375.774353027344	97.6896209716797\\
}--cycle;

\addplot[area legend, line width=0.5pt, draw=myblue, fill=myblue, fill opacity=0.3, forget plot]
table[row sep=crcr] {%
x	y\\
383.276123046875	231.818649291992\\
382.496490478516	237.612518310547\\
386.723876953125	238.181350708008\\
387.503509521484	232.387481689453\\
383.276123046875	231.818649291992\\
}--cycle;

\addplot[area legend, line width=0.5pt, draw=myblue, fill=myblue, fill opacity=0.3, forget plot]
table[row sep=crcr] {%
x	y\\
418.495971679688	229.035842895508\\
417.468170166016	236.083953857422\\
423.504028320312	236.964157104492\\
424.531829833984	229.916046142578\\
418.495971679688	229.035842895508\\
}--cycle;

\addplot[area legend, line width=0.5pt, draw=myblue, fill=myblue, fill opacity=0.3, forget plot]
table[row sep=crcr] {%
x	y\\
436.337860107422	278.73046875\\
442.995849609375	277.244232177734\\
441.662139892578	271.26953125\\
435.004150390625	272.755767822266\\
436.337860107422	278.73046875\\
}--cycle;

\addplot[area legend, line width=0.5pt, draw=myblue, fill=myblue, fill opacity=0.3, forget plot]
table[row sep=crcr] {%
x	y\\
445.742919921875	665.901977539062\\
443.508605957031	660.081970214844\\
438.257080078125	662.098022460938\\
440.491394042969	667.918029785156\\
445.742919921875	665.901977539062\\
}--cycle;

\addplot[area legend, line width=0.5pt, draw=myblue, fill=myblue, fill opacity=0.3, forget plot]
table[row sep=crcr] {%
x	y\\
453.975158691406	572.508422851562\\
455.875457763672	588.683959960938\\
466.024841308594	587.491577148438\\
464.124542236328	571.316040039062\\
453.975158691406	572.508422851562\\
}--cycle;

\addplot[area legend, line width=0.5pt, draw=myblue, fill=myblue, fill opacity=0.3, forget plot]
table[row sep=crcr] {%
x	y\\
465.926116943359	241.83740234375\\
467.703491210938	212.876312255859\\
456.073883056641	212.16259765625\\
454.296508789062	241.123687744141\\
465.926116943359	241.83740234375\\
}--cycle;

\addplot[area legend, line width=0.5pt, draw=myblue, fill=myblue, fill opacity=0.3, forget plot]
table[row sep=crcr] {%
x	y\\
467.916778564453	312.026428222656\\
471.189025878906	285.335113525391\\
460.083221435547	283.973571777344\\
456.810974121094	310.664886474609\\
467.916778564453	312.026428222656\\
}--cycle;

\addplot[area legend, line width=0.5pt, draw=myblue, fill=myblue, fill opacity=0.3, forget plot]
table[row sep=crcr] {%
x	y\\
476.450378417969	469.320098876953\\
474.391174316406	443.804931640625\\
463.549621582031	444.679901123047\\
465.608825683594	470.195068359375\\
476.450378417969	469.320098876953\\
}--cycle;

\addplot[area legend, line width=0.5pt, draw=myblue, fill=myblue, fill opacity=0.3, forget plot]
table[row sep=crcr] {%
x	y\\
471.179718017578	369.040466308594\\
465.158813476562	376.195495605469\\
470.820281982422	380.959533691406\\
476.841186523438	373.804504394531\\
471.179718017578	369.040466308594\\
}--cycle;

\addplot[area legend, line width=0.5pt, draw=myblue, fill=myblue, fill opacity=0.3, forget plot]
table[row sep=crcr] {%
x	y\\
472.066986083984	639.060302734375\\
474.874298095703	648.944519042969\\
481.933013916016	646.939697265625\\
479.125701904297	637.055480957031\\
472.066986083984	639.060302734375\\
}--cycle;

\addplot[area legend, line width=0.5pt, draw=myblue, fill=myblue, fill opacity=0.3, forget plot]
table[row sep=crcr] {%
x	y\\
503.095458984375	591.741943359375\\
499.021911621094	560.669494628906\\
486.904541015625	562.258056640625\\
490.978088378906	593.330505371094\\
503.095458984375	591.741943359375\\
}--cycle;

\addplot[area legend, line width=0.5pt, draw=myblue, fill=myblue, fill opacity=0.3, forget plot]
table[row sep=crcr] {%
x	y\\
504.23828125	799.767639160156\\
505.031890869141	822.605224609375\\
515.76171875	822.232360839844\\
514.968139648438	799.394775390625\\
504.23828125	799.767639160156\\
}--cycle;

\addplot[area legend, line width=0.5pt, draw=myblue, fill=myblue, fill opacity=0.3, forget plot]
table[row sep=crcr] {%
x	y\\
537.4951171875	323.407257080078\\
544.018798828125	323.994323730469\\
544.5048828125	318.592742919922\\
537.981201171875	318.005676269531\\
537.4951171875	323.407257080078\\
}--cycle;

\addplot[area legend, line width=0.5pt, draw=myblue, fill=myblue, fill opacity=0.3, forget plot]
table[row sep=crcr] {%
x	y\\
628.73388671875	310.653747558594\\
626.294860839844	305.127014160156\\
621.26611328125	307.346252441406\\
623.705139160156	312.872985839844\\
628.73388671875	310.653747558594\\
}--cycle;

\addplot[area legend, line width=0.5pt, draw=myblue, fill=myblue, fill opacity=0.3, forget plot]
table[row sep=crcr] {%
x	y\\
674.395568847656	157.074249267578\\
674.936767578125	151.437774658203\\
669.604431152344	150.925750732422\\
669.063232421875	156.562225341797\\
674.395568847656	157.074249267578\\
}--cycle;

\addplot[area legend, line width=0.5pt, draw=myblue, fill=myblue, fill opacity=0.3, forget plot]
table[row sep=crcr] {%
x	y\\
683.769653320312	194.302474975586\\
685.314147949219	199.236450195312\\
690.230346679688	197.697525024414\\
688.685852050781	192.763549804688\\
683.769653320312	194.302474975586\\
}--cycle;

\addplot[area legend, line width=0.5pt, draw=myblue, fill=myblue, fill opacity=0.3, forget plot]
table[row sep=crcr] {%
x	y\\
691.164611816406	223.96549987793\\
692.658203125	229.221542358398\\
696.835388183594	228.03450012207\\
695.341796875	222.778457641602\\
691.164611816406	223.96549987793\\
}--cycle;
\addplot [color=mygreen,  line width=0.5pt, forget plot]
  table[row sep=crcr]{%
323.728820800781	300.863525390625\\
328.971466064453	300.128021240234\\
328.271179199219	295.136474609375\\
323.028533935547	295.871978759766\\
323.728820800781	300.863525390625\\
};
\addplot [color=mygreen,  line width=0.5pt, forget plot]
  table[row sep=crcr]{%
335.225616455078	337.341766357422\\
339.485687255859	338.700042724609\\
340.774383544922	334.658233642578\\
336.514312744141	333.299957275391\\
335.225616455078	337.341766357422\\
};
\addplot [color=mygreen,  line width=0.5pt, forget plot]
  table[row sep=crcr]{%
353.138031005859	306.223937988281\\
359.648986816406	305.295562744141\\
358.861968994141	299.776062011719\\
352.351013183594	300.704437255859\\
353.138031005859	306.223937988281\\
};
\addplot [color=mygreen,  line width=0.5pt, forget plot]
  table[row sep=crcr]{%
371.478546142578	502.821136474609\\
365.510589599609	503.981994628906\\
366.521453857422	509.178863525391\\
372.489410400391	508.018005371094\\
371.478546142578	502.821136474609\\
};
\addplot [color=mygreen,  line width=0.5pt, forget plot]
  table[row sep=crcr]{%
373.875640869141	97.5765151977539\\
374.106201171875	102.69457244873\\
380.124359130859	102.423484802246\\
379.893798828125	97.3054275512695\\
373.875640869141	97.5765151977539\\
};
\addplot [color=mygreen,  line width=0.5pt, forget plot]
  table[row sep=crcr]{%
383.779174804688	230.356597900391\\
380.383056640625	234.490814208984\\
384.220825195312	237.643402099609\\
387.616943359375	233.509185791016\\
383.779174804688	230.356597900391\\
};
\addplot [color=mygreen,  line width=0.5pt, forget plot]
  table[row sep=crcr]{%
422.031158447266	229.105972290039\\
417.154205322266	231.801467895508\\
419.968841552734	236.894027709961\\
424.845794677734	234.198532104492\\
422.031158447266	229.105972290039\\
};
\addplot [color=mygreen,  line width=0.5pt, forget plot]
  table[row sep=crcr]{%
445.201568603516	666.566284179688\\
444.062774658203	660.453063964844\\
438.798431396484	661.433715820312\\
439.937225341797	667.546936035156\\
445.201568603516	666.566284179688\\
};
\addplot [color=mygreen,  line width=0.5pt, forget plot]
  table[row sep=crcr]{%
462.942535400391	235.477737426758\\
462.265197753906	220.202651977539\\
455.057464599609	220.522262573242\\
455.734802246094	235.797348022461\\
462.942535400391	235.477737426758\\
};
\addplot [color=mygreen,  line width=0.5pt, forget plot]
  table[row sep=crcr]{%
468.610107421875	307.808990478516\\
466.385498046875	283.197143554688\\
455.389892578125	284.191009521484\\
457.614501953125	308.802856445312\\
468.610107421875	307.808990478516\\
};
\addplot [color=mygreen,  line width=0.5pt, forget plot]
  table[row sep=crcr]{%
453.807403564453	576.043823242188\\
459.406524658203	598.630065917969\\
470.192596435547	595.956176757812\\
464.593475341797	573.369934082031\\
453.807403564453	576.043823242188\\
};
\addplot [color=mygreen,  line width=0.5pt, forget plot]
  table[row sep=crcr]{%
477.401702880859	467.696350097656\\
472.279113769531	440.1328125\\
460.598297119141	442.303649902344\\
465.720886230469	469.8671875\\
477.401702880859	467.696350097656\\
};
\addplot [color=mygreen,  line width=0.5pt, forget plot]
  table[row sep=crcr]{%
463.233062744141	375.806671142578\\
475.434692382812	378.418090820312\\
476.766937255859	372.193328857422\\
464.565307617188	369.581909179688\\
463.233062744141	375.806671142578\\
};
\addplot [color=mygreen,  line width=0.5pt, forget plot]
  table[row sep=crcr]{%
474.912231445312	640.987426757812\\
476.122650146484	646.171752929688\\
481.087768554688	645.012573242188\\
479.877349853516	639.828247070312\\
474.912231445312	640.987426757812\\
};
\addplot [color=mygreen,  line width=0.5pt, forget plot]
  table[row sep=crcr]{%
492.392303466797	808.481506347656\\
491.688110351562	804.191955566406\\
483.607696533203	805.518493652344\\
484.311889648438	809.808044433594\\
492.392303466797	808.481506347656\\
};
\addplot [color=mygreen,  line width=0.5pt, forget plot]
  table[row sep=crcr]{%
507.536865234375	593.972229003906\\
500.984710693359	563.348693847656\\
488.463134765625	566.027770996094\\
495.015289306641	596.651306152344\\
507.536865234375	593.972229003906\\
};
\addplot [color=mygreen,  line width=0.5pt, forget plot]
  table[row sep=crcr]{%
507.837982177734	800.207153320312\\
513.155395507812	827.905822753906\\
524.161987304688	825.792846679688\\
518.844604492188	798.094177246094\\
507.837982177734	800.207153320312\\
};
\addplot [color=mygreen,  line width=0.5pt, forget plot]
  table[row sep=crcr]{%
626.66748046875	309.383087158203\\
622.534973144531	306.310180664062\\
619.33251953125	310.616912841797\\
623.465026855469	313.689819335938\\
626.66748046875	309.383087158203\\
};
\addplot [color=myblue]
  table[row sep=crcr]{%
0	0\\
};
\addlegendentry{Netout}

\addplot [color=mygreen]
  table[row sep=crcr]{%
0	0\\
};
\addlegendentry{Label}

\end{axis}

\begin{axis}[%
width=\imagesize cm,
height=\imagesize cm,
at={(\offone cm, \offzero cm)},
scale only axis,
axis on top,
xmin=280.134128719563,
xmax=619.234474740324,
tick align=outside,
y dir=reverse,
ymin=163.759378798304,
ymax=502.859724819065,
axis line style={draw=none},
ticks=none,
title style={font=\bfseries},
title={Time step 2},
axis x line*=bottom,
axis y line*=left
]
\addplot [forget plot] graphics [xmin=0.5, xmax=901.5, ymin=0.5, ymax=901.5] {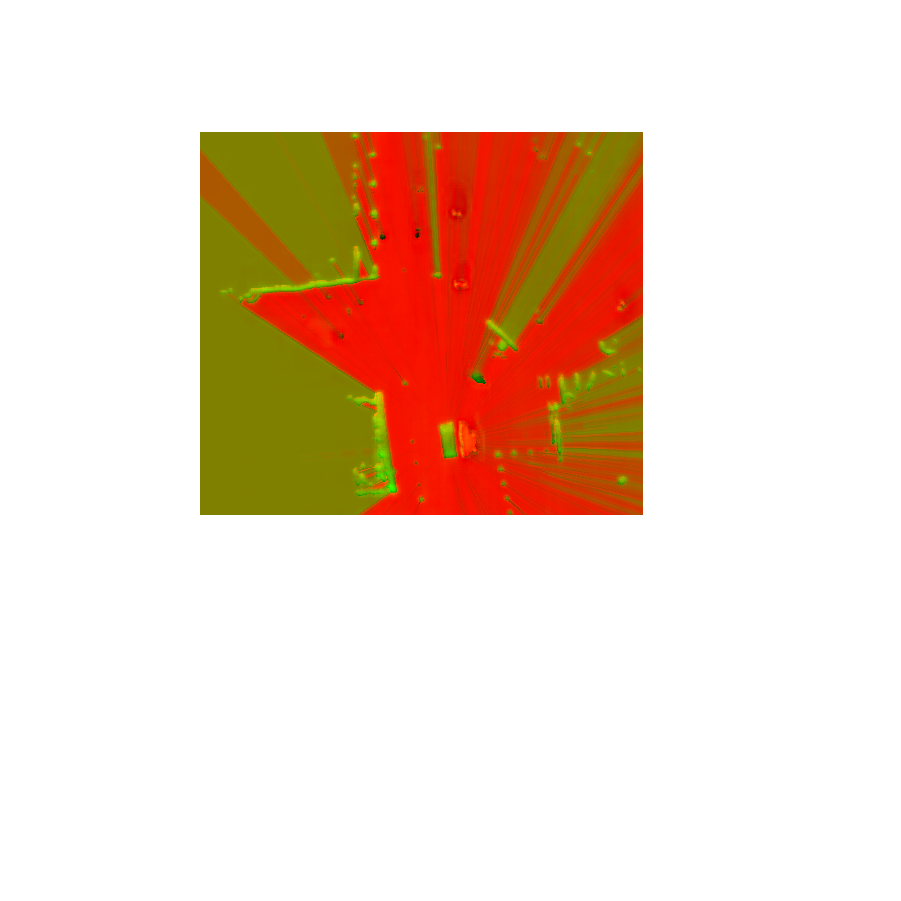};

\addplot[area legend, line width=0.5pt, draw=myblue, fill=myblue, fill opacity=0.3, forget plot]
table[row sep=crcr] {%
x	y\\
328.156097412109	299.283721923828\\
333.543914794922	297.274139404297\\
331.843902587891	292.716278076172\\
326.456085205078	294.725860595703\\
328.156097412109	299.283721923828\\
}--cycle;

\addplot[area legend, line width=0.5pt, draw=myblue, fill=myblue, fill opacity=0.3, forget plot]
table[row sep=crcr] {%
x	y\\
346.986907958984	334.060089111328\\
339.784942626953	332.94677734375\\
339.013092041016	337.939910888672\\
346.215057373047	339.05322265625\\
346.986907958984	334.060089111328\\
}--cycle;

\addplot[area legend, line width=0.5pt, draw=myblue, fill=myblue, fill opacity=0.3, forget plot]
table[row sep=crcr] {%
x	y\\
360.945159912109	305.710388183594\\
365.843231201172	302.330657958984\\
363.054840087891	298.289611816406\\
358.156768798828	301.669342041016\\
360.945159912109	305.710388183594\\
}--cycle;

\addplot[area legend, line width=0.5pt, draw=myblue, fill=myblue, fill opacity=0.3, forget plot]
table[row sep=crcr] {%
x	y\\
374.958953857422	101.853355407715\\
375.913726806641	107.082778930664\\
381.041046142578	106.146644592285\\
380.086273193359	100.917221069336\\
374.958953857422	101.853355407715\\
}--cycle;

\addplot[area legend, line width=0.5pt, draw=myblue, fill=myblue, fill opacity=0.3, forget plot]
table[row sep=crcr] {%
x	y\\
381.722290039062	234.174285888672\\
381.595153808594	239.718322753906\\
386.277709960938	239.825714111328\\
386.404846191406	234.281677246094\\
381.722290039062	234.174285888672\\
}--cycle;

\addplot[area legend, line width=0.5pt, draw=myblue, fill=myblue, fill opacity=0.3, forget plot]
table[row sep=crcr] {%
x	y\\
415.627288818359	230.792388916016\\
413.145233154297	237.17463684082\\
418.372711181641	239.207611083984\\
420.854766845703	232.82536315918\\
415.627288818359	230.792388916016\\
}--cycle;

\addplot[area legend, line width=0.5pt, draw=myblue, fill=myblue, fill opacity=0.3, forget plot]
table[row sep=crcr] {%
x	y\\
444.558380126953	663.431030273438\\
443.397125244141	657.418701171875\\
437.441619873047	658.568969726562\\
438.602874755859	664.581298828125\\
444.558380126953	663.431030273438\\
}--cycle;

\addplot[area legend, line width=0.5pt, draw=myblue, fill=myblue, fill opacity=0.3, forget plot]
table[row sep=crcr] {%
x	y\\
464.951629638672	219.889923095703\\
468.391723632812	191.483795166016\\
457.048370361328	190.110076904297\\
453.608276367188	218.516204833984\\
464.951629638672	219.889923095703\\
}--cycle;

\addplot[area legend, line width=0.5pt, draw=myblue, fill=myblue, fill opacity=0.3, forget plot]
table[row sep=crcr] {%
x	y\\
467.519104003906	287.694274902344\\
471.112365722656	261.779907226562\\
460.480895996094	260.305725097656\\
456.887634277344	286.220092773438\\
467.519104003906	287.694274902344\\
}--cycle;

\addplot[area legend, line width=0.5pt, draw=myblue, fill=myblue, fill opacity=0.3, forget plot]
table[row sep=crcr] {%
x	y\\
458.789581298828	586.759216308594\\
459.631652832031	601.77783203125\\
469.210418701172	601.240783691406\\
468.368347167969	586.22216796875\\
458.789581298828	586.759216308594\\
}--cycle;

\addplot[area legend, line width=0.5pt, draw=myblue, fill=myblue, fill opacity=0.3, forget plot]
table[row sep=crcr] {%
x	y\\
474.800476074219	452.580383300781\\
472.1845703125	426.325134277344\\
461.199523925781	427.419616699219\\
463.8154296875	453.674865722656\\
474.800476074219	452.580383300781\\
}--cycle;

\addplot[area legend, line width=0.5pt, draw=myblue, fill=myblue, fill opacity=0.3, forget plot]
table[row sep=crcr] {%
x	y\\
477.239929199219	374.433288574219\\
474.504028320312	380.933715820312\\
480.760070800781	383.566711425781\\
483.495971679688	377.066284179688\\
477.239929199219	374.433288574219\\
}--cycle;

\addplot[area legend, line width=0.5pt, draw=myblue, fill=myblue, fill opacity=0.3, forget plot]
table[row sep=crcr] {%
x	y\\
477.011047363281	653.4091796875\\
479.784454345703	664.407531738281\\
486.988952636719	662.5908203125\\
484.215545654297	651.592468261719\\
477.011047363281	653.4091796875\\
}--cycle;

\addplot[area legend, line width=0.5pt, draw=myblue, fill=myblue, fill opacity=0.3, forget plot]
table[row sep=crcr] {%
x	y\\
497.706909179688	573.686828613281\\
494.268981933594	542.97265625\\
482.293090820312	544.313171386719\\
485.731018066406	575.02734375\\
497.706909179688	573.686828613281\\
}--cycle;

\addplot[area legend, line width=0.5pt, draw=myblue, fill=myblue, fill opacity=0.3, forget plot]
table[row sep=crcr] {%
x	y\\
509.664825439453	811.520263671875\\
511.094421386719	837.107727050781\\
522.335144042969	836.479736328125\\
520.905578613281	810.892272949219\\
509.664825439453	811.520263671875\\
}--cycle;

\addplot[area legend, line width=0.5pt, draw=myblue, fill=myblue, fill opacity=0.3, forget plot]
table[row sep=crcr] {%
x	y\\
537.627319335938	318.594604492188\\
537.217346191406	325.079467773438\\
542.372680664062	325.405395507812\\
542.782653808594	318.920532226562\\
537.627319335938	318.594604492188\\
}--cycle;

\addplot[area legend, line width=0.5pt, draw=myblue, fill=myblue, fill opacity=0.3, forget plot]
table[row sep=crcr] {%
x	y\\
627.7919921875	305.892944335938\\
625.372192382812	299.990081787109\\
620.2080078125	302.107055664062\\
622.627807617188	308.009918212891\\
627.7919921875	305.892944335938\\
}--cycle;

\addplot[area legend, line width=0.5pt, draw=myblue, fill=myblue, fill opacity=0.3, forget plot]
table[row sep=crcr] {%
x	y\\
677.197143554688	157.269912719727\\
678.043212890625	151.498245239258\\
672.802856445312	150.730087280273\\
671.956787109375	156.501754760742\\
677.197143554688	157.269912719727\\
}--cycle;

\addplot[area legend, line width=0.5pt, draw=myblue, fill=myblue, fill opacity=0.3, forget plot]
table[row sep=crcr] {%
x	y\\
683.671264648438	197.231781005859\\
685.123962402344	202.269149780273\\
690.328735351562	200.768218994141\\
688.876037597656	195.730850219727\\
683.671264648438	197.231781005859\\
}--cycle;

\addplot[area legend, line width=0.5pt, draw=myblue, fill=myblue, fill opacity=0.3, forget plot]
table[row sep=crcr] {%
x	y\\
691.020324707031	227.075073242188\\
692.423522949219	232.177825927734\\
696.979675292969	230.924926757812\\
695.576477050781	225.822174072266\\
691.020324707031	227.075073242188\\
}--cycle;

\addplot[area legend, line width=0.5pt, draw=myblue, fill=myblue, fill opacity=0.3, forget plot]
table[row sep=crcr] {%
x	y\\
692.953491210938	199.858581542969\\
693.986450195312	205.13249206543\\
699.046508789062	204.141418457031\\
698.013549804688	198.86750793457\\
692.953491210938	199.858581542969\\
}--cycle;
\addplot [color=mygreen,  line width=0.5pt, forget plot]
  table[row sep=crcr]{%
327.803314208984	299.921081542969\\
333.025299072266	299.05078125\\
332.196685791016	294.078918457031\\
326.974700927734	294.94921875\\
327.803314208984	299.921081542969\\
};
\addplot [color=mygreen,  line width=0.5pt, forget plot]
  table[row sep=crcr]{%
339.577911376953	337.905517578125\\
344.030151367188	338.318603515625\\
344.422088623047	334.094482421875\\
339.969848632812	333.681396484375\\
339.577911376953	337.905517578125\\
};
\addplot [color=mygreen,  line width=0.5pt, forget plot]
  table[row sep=crcr]{%
356.981048583984	305.077423095703\\
363.530151367188	304.474395751953\\
363.018951416016	298.922576904297\\
356.469848632812	299.525604248047\\
356.981048583984	305.077423095703\\
};
\addplot [color=mygreen,  line width=0.5pt, forget plot]
  table[row sep=crcr]{%
374.909332275391	102.533714294434\\
375.069366455078	107.654457092285\\
381.090667724609	107.466285705566\\
380.930633544922	102.345542907715\\
374.909332275391	102.533714294434\\
};
\addplot [color=mygreen,  line width=0.5pt, forget plot]
  table[row sep=crcr]{%
382.627166748047	233.369003295898\\
379.40673828125	237.641479492188\\
383.372833251953	240.630996704102\\
386.59326171875	236.358520507812\\
382.627166748047	233.369003295898\\
};
\addplot [color=mygreen,  line width=0.5pt, forget plot]
  table[row sep=crcr]{%
418.82861328125	231.057907104492\\
414.097412109375	234.001754760742\\
417.17138671875	238.942092895508\\
421.902587890625	235.998245239258\\
418.82861328125	231.057907104492\\
};
\addplot [color=mygreen,  line width=0.5pt, forget plot]
  table[row sep=crcr]{%
445.072723388672	661.71923828125\\
444.233154296875	655.557800292969\\
438.927276611328	656.28076171875\\
439.766845703125	662.442199707031\\
445.072723388672	661.71923828125\\
};
\addplot [color=mygreen,  line width=0.5pt, forget plot]
  table[row sep=crcr]{%
463.589172363281	215.653625488281\\
463.625610351562	200.363571166992\\
456.410827636719	200.346374511719\\
456.374389648438	215.636428833008\\
463.589172363281	215.653625488281\\
};
\addplot [color=mygreen,  line width=0.5pt, forget plot]
  table[row sep=crcr]{%
468.387908935547	283.930633544922\\
466.624389648438	259.281494140625\\
455.612091064453	260.069366455078\\
457.375610351562	284.718505859375\\
468.387908935547	283.930633544922\\
};
\addplot [color=mygreen,  line width=0.5pt, forget plot]
  table[row sep=crcr]{%
456.923675537109	588.949279785156\\
462.259918212891	611.599060058594\\
473.076324462891	609.050720214844\\
467.740081787109	586.400939941406\\
456.923675537109	588.949279785156\\
};
\addplot [color=mygreen,  line width=0.5pt, forget plot]
  table[row sep=crcr]{%
476.411682128906	453.689727783203\\
471.267456054688	426.130249023438\\
459.588317871094	428.310272216797\\
464.732543945312	455.869750976562\\
476.411682128906	453.689727783203\\
};
\addplot [color=mygreen,  line width=0.5pt, forget plot]
  table[row sep=crcr]{%
472.127105712891	380.3486328125\\
484.126068115234	383.772735595703\\
485.872894287109	377.6513671875\\
473.873931884766	374.227264404297\\
472.127105712891	380.3486328125\\
};
\addplot [color=mygreen,  line width=0.5pt, forget plot]
  table[row sep=crcr]{%
477.928466796875	656.962768554688\\
479.097259521484	662.156616210938\\
484.071533203125	661.037231445312\\
482.902740478516	655.843383789062\\
477.928466796875	656.962768554688\\
};
\addplot [color=mygreen,  line width=0.5pt, forget plot]
  table[row sep=crcr]{%
493.401641845703	793.453552246094\\
492.670166015625	789.168579101562\\
484.598358154297	790.546447753906\\
485.329833984375	794.831420898438\\
493.401641845703	793.453552246094\\
};
\addplot [color=mygreen,  line width=0.5pt, forget plot]
  table[row sep=crcr]{%
501.886138916016	574.727294921875\\
494.563873291016	544.278747558594\\
482.113861083984	547.272705078125\\
489.436126708984	577.721252441406\\
501.886138916016	574.727294921875\\
};
\addplot [color=mygreen,  line width=0.5pt, forget plot]
  table[row sep=crcr]{%
503.690216064453	813.302673339844\\
509.328430175781	840.937805175781\\
520.309814453125	838.697326660156\\
514.671569824219	811.062194824219\\
503.690216064453	813.302673339844\\
};
\addplot [color=mygreen,  line width=0.5pt, forget plot]
  table[row sep=crcr]{%
624.607849121094	307.513488769531\\
620.827941894531	304.363555908203\\
617.392150878906	308.486511230469\\
621.172058105469	311.636444091797\\
624.607849121094	307.513488769531\\
};
\end{axis}

\begin{axis}[%
width=\imagesize cm,
height=\imagesize cm,
at={(\offtwo cm, \offzero cm)},
scale only axis,
axis on top,
xmin=280.134128719563,
xmax=619.234474740324,
tick align=outside,
y dir=reverse,
ymin=163.759378798304,
ymax=502.859724819065,
axis line style={draw=none},
ticks=none,
title style={font=\bfseries},
title={Time step 3},
axis x line*=bottom,
axis y line*=left
]
\addplot [forget plot] graphics [xmin=0.5, xmax=901.5, ymin=0.5, ymax=901.5] {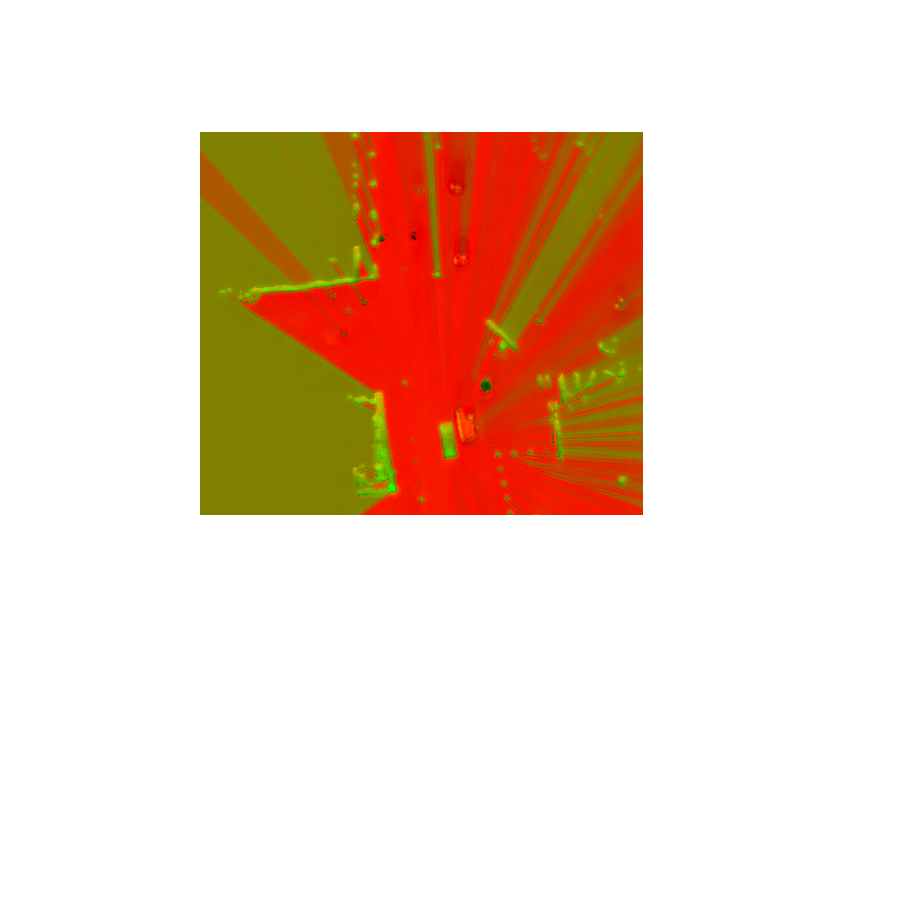};

\addplot[area legend, line width=0.5pt, draw=myblue, fill=myblue, fill opacity=0.3, forget plot]
table[row sep=crcr] {%
x	y\\
332.630676269531	299.788391113281\\
337.999847412109	296.477691650391\\
335.369323730469	292.211608886719\\
330.000152587891	295.522308349609\\
332.630676269531	299.788391113281\\
}--cycle;

\addplot[area legend, line width=0.5pt, draw=myblue, fill=myblue, fill opacity=0.3, forget plot]
table[row sep=crcr] {%
x	y\\
341.858306884766	336.921356201172\\
348.627716064453	336.290008544922\\
348.141693115234	331.078643798828\\
341.372283935547	331.709991455078\\
341.858306884766	336.921356201172\\
}--cycle;

\addplot[area legend, line width=0.5pt, draw=myblue, fill=myblue, fill opacity=0.3, forget plot]
table[row sep=crcr] {%
x	y\\
363.413146972656	305.6474609375\\
368.895599365234	302.803863525391\\
366.586853027344	298.3525390625\\
361.104400634766	301.196136474609\\
363.413146972656	305.6474609375\\
}--cycle;

\addplot[area legend, line width=0.5pt, draw=myblue, fill=myblue, fill opacity=0.3, forget plot]
table[row sep=crcr] {%
x	y\\
375.465667724609	106.296882629395\\
375.478942871094	111.71549987793\\
380.534332275391	111.703117370605\\
380.521057128906	106.28450012207\\
375.465667724609	106.296882629395\\
}--cycle;

\addplot[area legend, line width=0.5pt, draw=myblue, fill=myblue, fill opacity=0.3, forget plot]
table[row sep=crcr] {%
x	y\\
379.286712646484	236.141906738281\\
379.524536132812	242.06640625\\
384.713287353516	241.858093261719\\
384.475463867188	235.93359375\\
379.286712646484	236.141906738281\\
}--cycle;

\addplot[area legend, line width=0.5pt, draw=myblue, fill=myblue, fill opacity=0.3, forget plot]
table[row sep=crcr] {%
x	y\\
412.872802734375	233.037506103516\\
410.297149658203	238.805755615234\\
415.127197265625	240.962493896484\\
417.702850341797	235.194244384766\\
412.872802734375	233.037506103516\\
}--cycle;

\addplot[area legend, line width=0.5pt, draw=myblue, fill=myblue, fill opacity=0.3, forget plot]
table[row sep=crcr] {%
x	y\\
443.787109375	658.755798339844\\
441.674011230469	653.176025390625\\
436.212890625	655.244201660156\\
438.325988769531	660.823974609375\\
443.787109375	658.755798339844\\
}--cycle;

\addplot[area legend, line width=0.5pt, draw=myblue, fill=myblue, fill opacity=0.3, forget plot]
table[row sep=crcr] {%
x	y\\
463.674377441406	192.661529541016\\
464.005615234375	163.471008300781\\
452.325622558594	163.338470458984\\
451.994384765625	192.528991699219\\
463.674377441406	192.661529541016\\
}--cycle;

\addplot[area legend, line width=0.5pt, draw=myblue, fill=myblue, fill opacity=0.3, forget plot]
table[row sep=crcr] {%
x	y\\
468.446075439453	264.193145751953\\
470.767669677734	236.755706787109\\
459.553924560547	235.806869506836\\
457.232330322266	263.244293212891\\
468.446075439453	264.193145751953\\
}--cycle;

\addplot[area legend, line width=0.5pt, draw=myblue, fill=myblue, fill opacity=0.3, forget plot]
table[row sep=crcr] {%
x	y\\
472.461853027344	438.696990966797\\
470.532257080078	412.493408203125\\
459.538146972656	413.303009033203\\
461.467742919922	439.506591796875\\
472.461853027344	438.696990966797\\
}--cycle;

\addplot[area legend, line width=0.5pt, draw=myblue, fill=myblue, fill opacity=0.3, forget plot]
table[row sep=crcr] {%
x	y\\
464.361602783203	601.26123046875\\
464.128509521484	616.59423828125\\
473.638397216797	616.73876953125\\
473.871490478516	601.40576171875\\
464.361602783203	601.26123046875\\
}--cycle;

\addplot[area legend, line width=0.5pt, draw=myblue, fill=myblue, fill opacity=0.3, forget plot]
table[row sep=crcr] {%
x	y\\
492.235626220703	553.911193847656\\
490.304351806641	521.345092773438\\
477.764373779297	522.088806152344\\
479.695648193359	554.654907226562\\
492.235626220703	553.911193847656\\
}--cycle;

\addplot[area legend, line width=0.5pt, draw=myblue, fill=myblue, fill opacity=0.3, forget plot]
table[row sep=crcr] {%
x	y\\
483.606475830078	382.832244873047\\
481.705780029297	389.158386230469\\
488.393524169922	391.167755126953\\
490.294219970703	384.841613769531\\
483.606475830078	382.832244873047\\
}--cycle;

\addplot[area legend, line width=0.5pt, draw=myblue, fill=myblue, fill opacity=0.3, forget plot]
table[row sep=crcr] {%
x	y\\
508.796264648438	622.35205078125\\
509.366790771484	591.875671386719\\
497.203735351562	591.64794921875\\
496.633209228516	622.124328613281\\
508.796264648438	622.35205078125\\
}--cycle;

\addplot[area legend, line width=0.5pt, draw=myblue, fill=myblue, fill opacity=0.3, forget plot]
table[row sep=crcr] {%
x	y\\
511.434692382812	823.102416992188\\
512.091735839844	845.208862304688\\
522.565307617188	844.897583007812\\
521.908264160156	822.791137695312\\
511.434692382812	823.102416992188\\
}--cycle;

\addplot[area legend, line width=0.5pt, draw=myblue, fill=myblue, fill opacity=0.3, forget plot]
table[row sep=crcr] {%
x	y\\
537.289367675781	318.796752929688\\
537.367614746094	325.267852783203\\
542.710632324219	325.203247070312\\
542.632385253906	318.732147216797\\
537.289367675781	318.796752929688\\
}--cycle;

\addplot[area legend, line width=0.5pt, draw=myblue, fill=myblue, fill opacity=0.3, forget plot]
table[row sep=crcr] {%
x	y\\
622.076782226562	310.070922851562\\
621.705749511719	303.59765625\\
615.923217773438	303.929077148438\\
616.294250488281	310.40234375\\
622.076782226562	310.070922851562\\
}--cycle;

\addplot[area legend, line width=0.5pt, draw=myblue, fill=myblue, fill opacity=0.3, forget plot]
table[row sep=crcr] {%
x	y\\
684.638000488281	202.331268310547\\
686.514770507812	207.447021484375\\
691.361999511719	205.668731689453\\
689.485229492188	200.552978515625\\
684.638000488281	202.331268310547\\
}--cycle;

\addplot[area legend, line width=0.5pt, draw=myblue, fill=myblue, fill opacity=0.3, forget plot]
table[row sep=crcr] {%
x	y\\
696.498962402344	188.059463500977\\
696.888000488281	182.304702758789\\
691.501037597656	181.940536499023\\
691.111999511719	187.695297241211\\
696.498962402344	188.059463500977\\
}--cycle;
\addplot [color=mygreen,  line width=0.5pt, forget plot]
  table[row sep=crcr]{%
331.812896728516	299.928253173828\\
337.031982421875	299.040863037109\\
336.187103271484	294.071746826172\\
330.968017578125	294.959136962891\\
331.812896728516	299.928253173828\\
};
\addplot [color=mygreen,  line width=0.5pt, forget plot]
  table[row sep=crcr]{%
343.984130859375	336.3310546875\\
348.433746337891	335.890594482422\\
348.015869140625	331.6689453125\\
343.566253662109	332.109405517578\\
343.984130859375	336.3310546875\\
};
\addplot [color=mygreen,  line width=0.5pt, forget plot]
  table[row sep=crcr]{%
361.640045166016	304.700988769531\\
368.214599609375	304.872436523438\\
368.359954833984	299.299011230469\\
361.785400390625	299.127563476562\\
361.640045166016	304.700988769531\\
};
\addplot [color=mygreen,  line width=0.5pt, forget plot]
  table[row sep=crcr]{%
373.902801513672	105.541893005371\\
374.076385498047	110.662200927734\\
380.097198486328	110.458106994629\\
379.923614501953	105.337799072266\\
373.902801513672	105.541893005371\\
};
\addplot [color=mygreen,  line width=0.5pt, forget plot]
  table[row sep=crcr]{%
381.194030761719	234.440017700195\\
378.509704589844	239.068161010742\\
382.805969238281	241.559982299805\\
385.490295410156	236.931838989258\\
381.194030761719	234.440017700195\\
};
\addplot [color=mygreen,  line width=0.5pt, forget plot]
  table[row sep=crcr]{%
413.933624267578	233.081451416016\\
409.125457763672	235.897872924805\\
412.066375732422	240.918548583984\\
416.874542236328	238.102127075195\\
413.933624267578	233.081451416016\\
};
\addplot [color=mygreen,  line width=0.5pt, forget plot]
  table[row sep=crcr]{%
444.083648681641	658.706787109375\\
443.21923828125	652.548828125\\
437.916351318359	653.293212890625\\
438.78076171875	659.451171875\\
444.083648681641	658.706787109375\\
};
\addplot [color=mygreen,  line width=0.5pt, forget plot]
  table[row sep=crcr]{%
462.673858642578	189.613342285156\\
462.540710449219	174.323837280273\\
455.326141357422	174.386657714844\\
455.459289550781	189.676162719727\\
462.673858642578	189.613342285156\\
};
\addplot [color=mygreen,  line width=0.5pt, forget plot]
  table[row sep=crcr]{%
470.010375976562	260.125213623047\\
469.021209716797	235.432861328125\\
457.989624023438	235.874771118164\\
458.978790283203	260.567138671875\\
470.010375976562	260.125213623047\\
};
\addplot [color=mygreen,  line width=0.5pt, forget plot]
  table[row sep=crcr]{%
475.125915527344	436.874572753906\\
470.598999023438	409.207000732422\\
458.874084472656	411.125427246094\\
463.401000976562	438.792999267578\\
475.125915527344	436.874572753906\\
};
\addplot [color=mygreen,  line width=0.5pt, forget plot]
  table[row sep=crcr]{%
459.833740234375	605.022216796875\\
465.373199462891	627.623168945312\\
476.166259765625	624.977783203125\\
470.626800537109	602.376831054688\\
459.833740234375	605.022216796875\\
};
\addplot [color=mygreen,  line width=0.5pt, forget plot]
  table[row sep=crcr]{%
480.913787841797	673.985046386719\\
482.120208740234	679.170349121094\\
487.086212158203	678.014953613281\\
485.879791259766	672.829650878906\\
480.913787841797	673.985046386719\\
};
\addplot [color=mygreen,  line width=0.5pt, forget plot]
  table[row sep=crcr]{%
489.389221191406	778.490600585938\\
488.693939208984	774.199645996094\\
480.610778808594	775.509399414062\\
481.306060791016	779.800354003906\\
489.389221191406	778.490600585938\\
};
\addplot [color=mygreen,  line width=0.5pt, forget plot]
  table[row sep=crcr]{%
480.001190185547	387.731079101562\\
490.890380859375	393.824127197266\\
493.998809814453	388.268920898438\\
483.109619140625	382.175872802734\\
480.001190185547	387.731079101562\\
};
\addplot [color=mygreen,  line width=0.5pt, forget plot]
  table[row sep=crcr]{%
496.848785400391	555.754089355469\\
489.609313964844	525.285766601562\\
477.151214599609	528.245910644531\\
484.390686035156	558.714233398438\\
496.848785400391	555.754089355469\\
};
\addplot [color=mygreen,  line width=0.5pt, forget plot]
  table[row sep=crcr]{%
504.889221191406	826.174621582031\\
510.095825195312	853.894287109375\\
521.110778808594	851.825378417969\\
515.904174804688	824.105712890625\\
504.889221191406	826.174621582031\\
};
\addplot [color=mygreen,  line width=0.5pt, forget plot]
  table[row sep=crcr]{%
621.598754882812	305.450439453125\\
617.764343261719	302.367126464844\\
614.401245117188	306.549560546875\\
618.235656738281	309.632873535156\\
621.598754882812	305.450439453125\\
};
\addplot [color=mygreen,  line width=0.5pt, forget plot]
  table[row sep=crcr]{%
627.1455078125	300.805480957031\\
624.338317871094	294.680999755859\\
618.8544921875	297.194519042969\\
621.661682128906	303.319000244141\\
627.1455078125	300.805480957031\\
};
\end{axis}

\begin{axis}[%
width=\imagesize cm,
height=\imagesize cm,
at={(\offthree cm, \offzero cm)},
scale only axis,
axis on top,
xmin=280.134128719563,
xmax=619.234474740324,
tick align=outside,
y dir=reverse,
ymin=163.759378798304,
ymax=502.859724819065,
axis line style={draw=none},
ticks=none,
title style={font=\bfseries},
title={Time step 4},
axis x line*=bottom,
axis y line*=left
]
\addplot [forget plot] graphics [xmin=0.5, xmax=901.5, ymin=0.5, ymax=901.5] {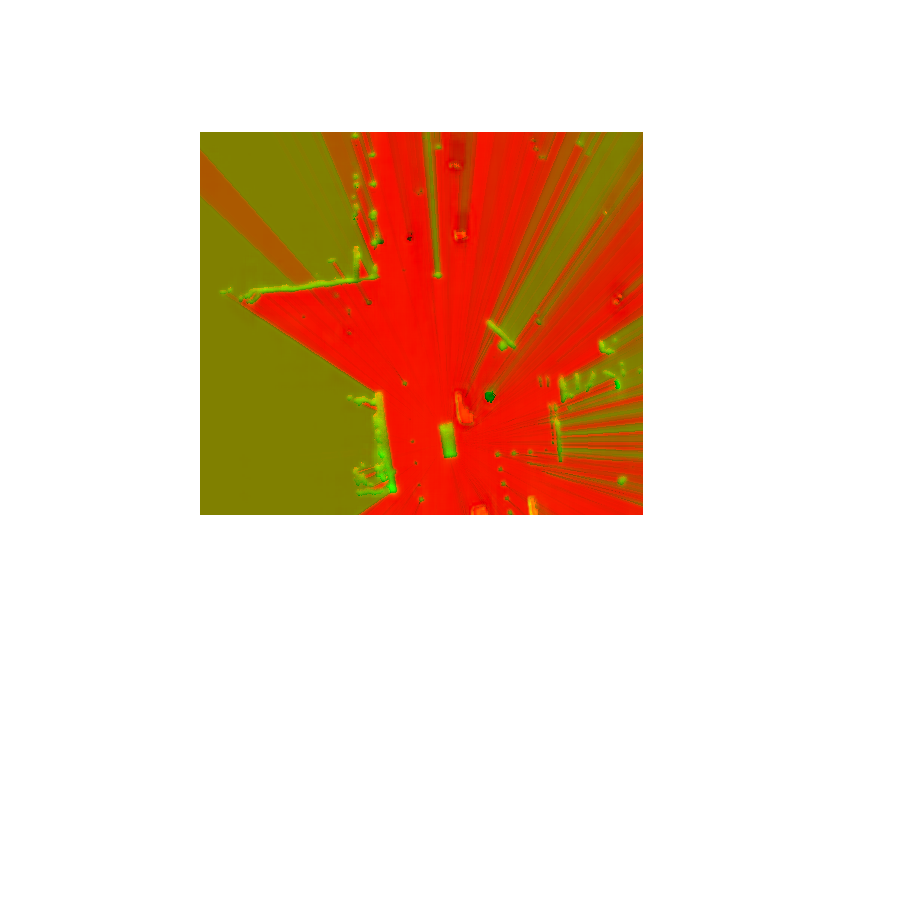};

\addplot[area legend, line width=0.5pt, draw=myblue, fill=myblue, fill opacity=0.3, forget plot]
table[row sep=crcr] {%
x	y\\
335.429779052734	298.387725830078\\
340.531768798828	296.211944580078\\
338.570220947266	291.612274169922\\
333.468231201172	293.788055419922\\
335.429779052734	298.387725830078\\
}--cycle;

\addplot[area legend, line width=0.5pt, draw=myblue, fill=myblue, fill opacity=0.3, forget plot]
table[row sep=crcr] {%
x	y\\
351.678344726562	336.444885253906\\
353.270721435547	331.003387451172\\
348.321655273438	329.555114746094\\
346.729278564453	334.996612548828\\
351.678344726562	336.444885253906\\
}--cycle;

\addplot[area legend, line width=0.5pt, draw=myblue, fill=myblue, fill opacity=0.3, forget plot]
table[row sep=crcr] {%
x	y\\
367.076354980469	305.407775878906\\
372.717620849609	303.221710205078\\
370.923645019531	298.592224121094\\
365.282379150391	300.778289794922\\
367.076354980469	305.407775878906\\
}--cycle;

\addplot[area legend, line width=0.5pt, draw=myblue, fill=myblue, fill opacity=0.3, forget plot]
table[row sep=crcr] {%
x	y\\
373.717742919922	111.142570495605\\
375.105529785156	116.261016845703\\
380.282257080078	114.857429504395\\
378.894470214844	109.738983154297\\
373.717742919922	111.142570495605\\
}--cycle;

\addplot[area legend, line width=0.5pt, draw=myblue, fill=myblue, fill opacity=0.3, forget plot]
table[row sep=crcr] {%
x	y\\
378.333892822266	238.105178833008\\
378.236206054688	243.801712036133\\
383.666107177734	243.894821166992\\
383.763793945312	238.198287963867\\
378.333892822266	238.105178833008\\
}--cycle;

\addplot[area legend, line width=0.5pt, draw=myblue, fill=myblue, fill opacity=0.3, forget plot]
table[row sep=crcr] {%
x	y\\
409.618469238281	234.132598876953\\
406.257049560547	239.045333862305\\
410.381530761719	241.867401123047\\
413.742950439453	236.954666137695\\
409.618469238281	234.132598876953\\
}--cycle;

\addplot[area legend, line width=0.5pt, draw=myblue, fill=myblue, fill opacity=0.3, forget plot]
table[row sep=crcr] {%
x	y\\
443.498016357422	654.244323730469\\
442.075103759766	648.399047851562\\
436.501983642578	649.755676269531\\
437.924896240234	655.600952148438\\
443.498016357422	654.244323730469\\
}--cycle;

\addplot[area legend, line width=0.5pt, draw=myblue, fill=myblue, fill opacity=0.3, forget plot]
table[row sep=crcr] {%
x	y\\
462.917602539062	170.677124023438\\
464.277160644531	143.891082763672\\
453.082397460938	143.322875976562\\
451.722839355469	170.108917236328\\
462.917602539062	170.677124023438\\
}--cycle;

\addplot[area legend, line width=0.5pt, draw=myblue, fill=myblue, fill opacity=0.3, forget plot]
table[row sep=crcr] {%
x	y\\
467.990814208984	240.530944824219\\
471.340637207031	212.83984375\\
460.009185791016	211.469055175781\\
456.659362792969	239.16015625\\
467.990814208984	240.530944824219\\
}--cycle;

\addplot[area legend, line width=0.5pt, draw=myblue, fill=myblue, fill opacity=0.3, forget plot]
table[row sep=crcr] {%
x	y\\
470.422393798828	417.716796875\\
468.993103027344	389.700805664062\\
457.577606201172	390.283203125\\
459.006896972656	418.299194335938\\
470.422393798828	417.716796875\\
}--cycle;

\addplot[area legend, line width=0.5pt, draw=myblue, fill=myblue, fill opacity=0.3, forget plot]
table[row sep=crcr] {%
x	y\\
465.785003662109	618.592224121094\\
466.583221435547	635.853210449219\\
476.214996337891	635.407775878906\\
475.416778564453	618.146789550781\\
465.785003662109	618.592224121094\\
}--cycle;

\addplot[area legend, line width=0.5pt, draw=myblue, fill=myblue, fill opacity=0.3, forget plot]
table[row sep=crcr] {%
x	y\\
488.512573242188	540.665283203125\\
487.109771728516	504.801849365234\\
473.487426757812	505.334686279297\\
474.890228271484	541.198120117188\\
488.512573242188	540.665283203125\\
}--cycle;

\addplot[area legend, line width=0.5pt, draw=myblue, fill=myblue, fill opacity=0.3, forget plot]
table[row sep=crcr] {%
x	y\\
488.5361328125	391.438934326172\\
484.699462890625	399.229919433594\\
491.4638671875	402.561065673828\\
495.300537109375	394.770080566406\\
488.5361328125	391.438934326172\\
}--cycle;

\addplot[area legend, line width=0.5pt, draw=myblue, fill=myblue, fill opacity=0.3, forget plot]
table[row sep=crcr] {%
x	y\\
484.693969726562	688.875\\
485.217681884766	709.382629394531\\
495.306030273438	709.125\\
494.782318115234	688.617370605469\\
484.693969726562	688.875\\
}--cycle;

\addplot[area legend, line width=0.5pt, draw=myblue, fill=myblue, fill opacity=0.3, forget plot]
table[row sep=crcr] {%
x	y\\
488.354156494141	759.527648925781\\
486.357788085938	785.608337402344\\
497.645843505859	786.472351074219\\
499.642211914062	760.391662597656\\
488.354156494141	759.527648925781\\
}--cycle;

\addplot[area legend, line width=0.5pt, draw=myblue, fill=myblue, fill opacity=0.3, forget plot]
table[row sep=crcr] {%
x	y\\
504.120941162109	601.092590332031\\
505.806945800781	571.589050292969\\
493.879058837891	570.907409667969\\
492.193054199219	600.410949707031\\
504.120941162109	601.092590332031\\
}--cycle;

\addplot[area legend, line width=0.5pt, draw=myblue, fill=myblue, fill opacity=0.3, forget plot]
table[row sep=crcr] {%
x	y\\
512.589965820312	835.199401855469\\
513.008666992188	857.000366210938\\
523.410034179688	856.800598144531\\
522.991333007812	834.999633789062\\
512.589965820312	835.199401855469\\
}--cycle;

\addplot[area legend, line width=0.5pt, draw=myblue, fill=myblue, fill opacity=0.3, forget plot]
table[row sep=crcr] {%
x	y\\
532.083374023438	141.91845703125\\
537.089782714844	140.626052856445\\
535.916625976562	136.08154296875\\
530.910217285156	137.373947143555\\
532.083374023438	141.91845703125\\
}--cycle;

\addplot[area legend, line width=0.5pt, draw=myblue, fill=myblue, fill opacity=0.3, forget plot]
table[row sep=crcr] {%
x	y\\
535.876098632812	319.132537841797\\
536.646423339844	325.52734375\\
542.123901367188	324.867462158203\\
541.353576660156	318.47265625\\
535.876098632812	319.132537841797\\
}--cycle;

\addplot[area legend, line width=0.5pt, draw=myblue, fill=myblue, fill opacity=0.3, forget plot]
table[row sep=crcr] {%
x	y\\
607.801574707031	217.876251220703\\
610.723999023438	212.658462524414\\
606.198425292969	210.123748779297\\
603.276000976562	215.341537475586\\
607.801574707031	217.876251220703\\
}--cycle;

\addplot[area legend, line width=0.5pt, draw=myblue, fill=myblue, fill opacity=0.3, forget plot]
table[row sep=crcr] {%
x	y\\
620.926208496094	303.968414306641\\
618.83447265625	298.009460449219\\
613.073791503906	300.031585693359\\
615.16552734375	305.990539550781\\
620.926208496094	303.968414306641\\
}--cycle;

\addplot[area legend, line width=0.5pt, draw=myblue, fill=myblue, fill opacity=0.3, forget plot]
table[row sep=crcr] {%
x	y\\
625.569152832031	299.321746826172\\
621.841674804688	293.318603515625\\
616.430847167969	296.678253173828\\
620.158325195312	302.681396484375\\
625.569152832031	299.321746826172\\
}--cycle;

\addplot[area legend, line width=0.5pt, draw=myblue, fill=myblue, fill opacity=0.3, forget plot]
table[row sep=crcr] {%
x	y\\
684.814208984375	205.923080444336\\
686.495239257812	211.492660522461\\
691.185791015625	210.076919555664\\
689.504760742188	204.507339477539\\
684.814208984375	205.923080444336\\
}--cycle;

\addplot[area legend, line width=0.5pt, draw=myblue, fill=myblue, fill opacity=0.3, forget plot]
table[row sep=crcr] {%
x	y\\
695.325866699219	183.425231933594\\
694.247924804688	177.551834106445\\
688.674133300781	178.574768066406\\
689.752075195312	184.448165893555\\
695.325866699219	183.425231933594\\
}--cycle;
\addplot [color=mygreen,  line width=0.5pt, forget plot]
  table[row sep=crcr]{%
335.030914306641	298.079071044922\\
340.171936035156	296.815734863281\\
338.969085693359	291.920928955078\\
333.828063964844	293.184265136719\\
335.030914306641	298.079071044922\\
};
\addplot [color=mygreen,  line width=0.5pt, forget plot]
  table[row sep=crcr]{%
348.376129150391	335.619262695312\\
352.700988769531	334.483978271484\\
351.623870849609	330.380737304688\\
347.299011230469	331.516021728516\\
348.376129150391	335.619262695312\\
};
\addplot [color=mygreen,  line width=0.5pt, forget plot]
  table[row sep=crcr]{%
365.802673339844	304.891662597656\\
372.376098632812	304.680786132812\\
372.197326660156	299.108337402344\\
365.623901367188	299.319213867188\\
365.802673339844	304.891662597656\\
};
\addplot [color=mygreen,  line width=0.5pt, forget plot]
  table[row sep=crcr]{%
373.106414794922	110.30517578125\\
372.875427246094	115.423210144043\\
378.893585205078	115.69482421875\\
379.124572753906	110.576789855957\\
373.106414794922	110.30517578125\\
};
\addplot [color=mygreen,  line width=0.5pt, forget plot]
  table[row sep=crcr]{%
379.640594482422	238.612503051758\\
377.722839355469	243.607238769531\\
382.359405517578	245.387496948242\\
384.277160644531	240.392761230469\\
379.640594482422	238.612503051758\\
};
\addplot [color=mygreen,  line width=0.5pt, forget plot]
  table[row sep=crcr]{%
410.016937255859	234.102233886719\\
405.149841308594	236.815536499023\\
407.983062744141	241.897766113281\\
412.850158691406	239.184463500977\\
410.016937255859	234.102233886719\\
};
\addplot [color=mygreen,  line width=0.5pt, forget plot]
  table[row sep=crcr]{%
443.139770507812	654.641479492188\\
442.146331787109	648.502990722656\\
436.860229492188	649.358520507812\\
437.853668212891	655.497009277344\\
443.139770507812	654.641479492188\\
};
\addplot [color=mygreen,  line width=0.5pt, forget plot]
  table[row sep=crcr]{%
459.663513183594	168.8720703125\\
458.954528808594	158.668106079102\\
452.336486816406	159.1279296875\\
453.045471191406	169.331893920898\\
459.663513183594	168.8720703125\\
};
\addplot [color=mygreen,  line width=0.5pt, forget plot]
  table[row sep=crcr]{%
467.725921630859	235.262100219727\\
467.312957763672	210.55339050293\\
456.274078369141	210.737899780273\\
456.687042236328	235.44660949707\\
467.725921630859	235.262100219727\\
};
\addplot [color=mygreen,  line width=0.5pt, forget plot]
  table[row sep=crcr]{%
472.75634765625	422.100555419922\\
469.018402099609	394.315368652344\\
457.24365234375	395.899444580078\\
460.981597900391	423.684631347656\\
472.75634765625	422.100555419922\\
};
\addplot [color=mygreen,  line width=0.5pt, forget plot]
  table[row sep=crcr]{%
465.682403564453	621.148010253906\\
471.566131591797	643.661804199219\\
482.317596435547	640.851989746094\\
476.433868408203	618.338195800781\\
465.682403564453	621.148010253906\\
};
\addplot [color=mygreen,  line width=0.5pt, forget plot]
  table[row sep=crcr]{%
492.634857177734	537.904846191406\\
485.867553710938	507.328125\\
473.365142822266	510.095153808594\\
480.132446289062	540.671875\\
492.634857177734	537.904846191406\\
};
\addplot [color=mygreen,  line width=0.5pt, forget plot]
  table[row sep=crcr]{%
488.386657714844	764.498168945312\\
487.69873046875	760.205993652344\\
479.613342285156	761.501831054688\\
480.30126953125	765.794006347656\\
488.386657714844	764.498168945312\\
};
\addplot [color=mygreen,  line width=0.5pt, forget plot]
  table[row sep=crcr]{%
484.878784179688	689.039733886719\\
486.176330566406	694.202941894531\\
491.121215820312	692.960266113281\\
489.823669433594	687.797058105469\\
484.878784179688	689.039733886719\\
};
\addplot [color=mygreen,  line width=0.5pt, forget plot]
  table[row sep=crcr]{%
483.633453369141	393.080627441406\\
491.373443603516	402.867980957031\\
496.366546630859	398.919372558594\\
488.626556396484	389.132019042969\\
483.633453369141	393.080627441406\\
};
\addplot [color=mygreen,  line width=0.5pt, forget plot]
  table[row sep=crcr]{%
506.358764648438	600.672424316406\\
500.688385009766	573.05908203125\\
489.641235351562	575.327575683594\\
495.311614990234	602.94091796875\\
506.358764648438	600.672424316406\\
};
\addplot [color=mygreen,  line width=0.5pt, forget plot]
  table[row sep=crcr]{%
507.189208984375	836.98974609375\\
511.750762939453	864.822875976562\\
522.810791015625	863.01025390625\\
518.249267578125	835.177124023438\\
507.189208984375	836.98974609375\\
};
\addplot [color=mygreen,  line width=0.5pt, forget plot]
  table[row sep=crcr]{%
620.640441894531	301.976470947266\\
617.292053222656	298.371246337891\\
613.359558105469	302.023529052734\\
616.707946777344	305.628753662109\\
620.640441894531	301.976470947266\\
};
\addplot [color=mygreen,  line width=0.5pt, forget plot]
  table[row sep=crcr]{%
625.834899902344	298.3955078125\\
623.958923339844	291.924774169922\\
618.165100097656	293.6044921875\\
620.041076660156	300.075225830078\\
625.834899902344	298.3955078125\\
};
\end{axis}
\end{tikzpicture}%

%% file: doc/sec_conclusion.tex
\mysection{Conclusion}{conclusions}
In this paper, we presented a new deep learning approach to obtain the size, position and orientation of dynamic objects, represented as bounding boxes, as well as a pixel-wise segmentation of dynamic and static cells. We build on our previous network structure and extended it with a convolutional LSTM cell. This way, we are able to use memory to track objects states over time and overcome the problem of short occlusions and gain an enhanced object detection performance. Furthermore, the LSTM cell improves the consistency of size and position estimates, because previous knowledge about the object states is used for sequential filtering. Besides the LSTM cell, we included the classification of static and dynamic cells into our environment model, thus increasing the information value of our representation.

In the future, we wish to use raw measurements as input instead of a dynamic occupancy grid map constructed through Bayesian sensor fusion, thus employing an end-to-end learning approach. Our experiment using only the masses for free and occupied space without the velocities showed promising object detection results, that we wish to improve upon. Furthermore, for future work, parametric free space should be learned by the network, which is currently not possible with the proposed network structure. Finally, the extension of our work to use data from a moving vehicle can be considered.